\documentclass[10pt,twocolumn,letterpaper]{article}

\usepackage[pagenumbers]{paperstyle}

\usepackage{cuted}   
\usepackage{caption} 

\definecolor{linkblue}{rgb}{0.21,0.49,0.74}
\usepackage[pagebackref,breaklinks,colorlinks,allcolors=linkblue]{hyperref}

\usepackage[table]{xcolor}
\usepackage{booktabs}
\usepackage{multirow}
\usepackage{graphicx}

\title{RoboCap: A New Platform for Egocentric Robot Learning}

\author{%
Qasim Ali\textsuperscript{1} \quad
Nikash Bhardwaj\textsuperscript{1} \quad
Kheri Hughes\textsuperscript{1} \quad
Sam Cho\textsuperscript{2} \quad
Ethan Lee\textsuperscript{2} \quad
Paul Luo\textsuperscript{2} \quad \\
Shinn Chong\textsuperscript{2} \quad
Sunny Yu\textsuperscript{2} \quad
Owen Akhibi\textsuperscript{1} \quad
Alex Proshkin\textsuperscript{1} \quad
Haotian Zhan\textsuperscript{1} \quad \\
Niresh Dravin\textsuperscript{2} \quad
Vivek Sharma\textsuperscript{2} \quad
Jonathan Victor\textsuperscript{2} \quad
Amanda Young\textsuperscript{2} \quad
Michael Cho\textsuperscript{2} \quad \\
Ademi Adeniji\textsuperscript{1} \quad
Vincent Liu\textsuperscript{1} \\[0.6em]
\textsuperscript{1}Grounded Superintelligence \quad
\textsuperscript{2}BitRobot \\[0.6em]
\href{https://www.gsi.company}{\texttt{gsi.company}}
}

\begin{document}
\maketitle
\begin{strip}
    \centering
    \includegraphics[width=0.32\linewidth]{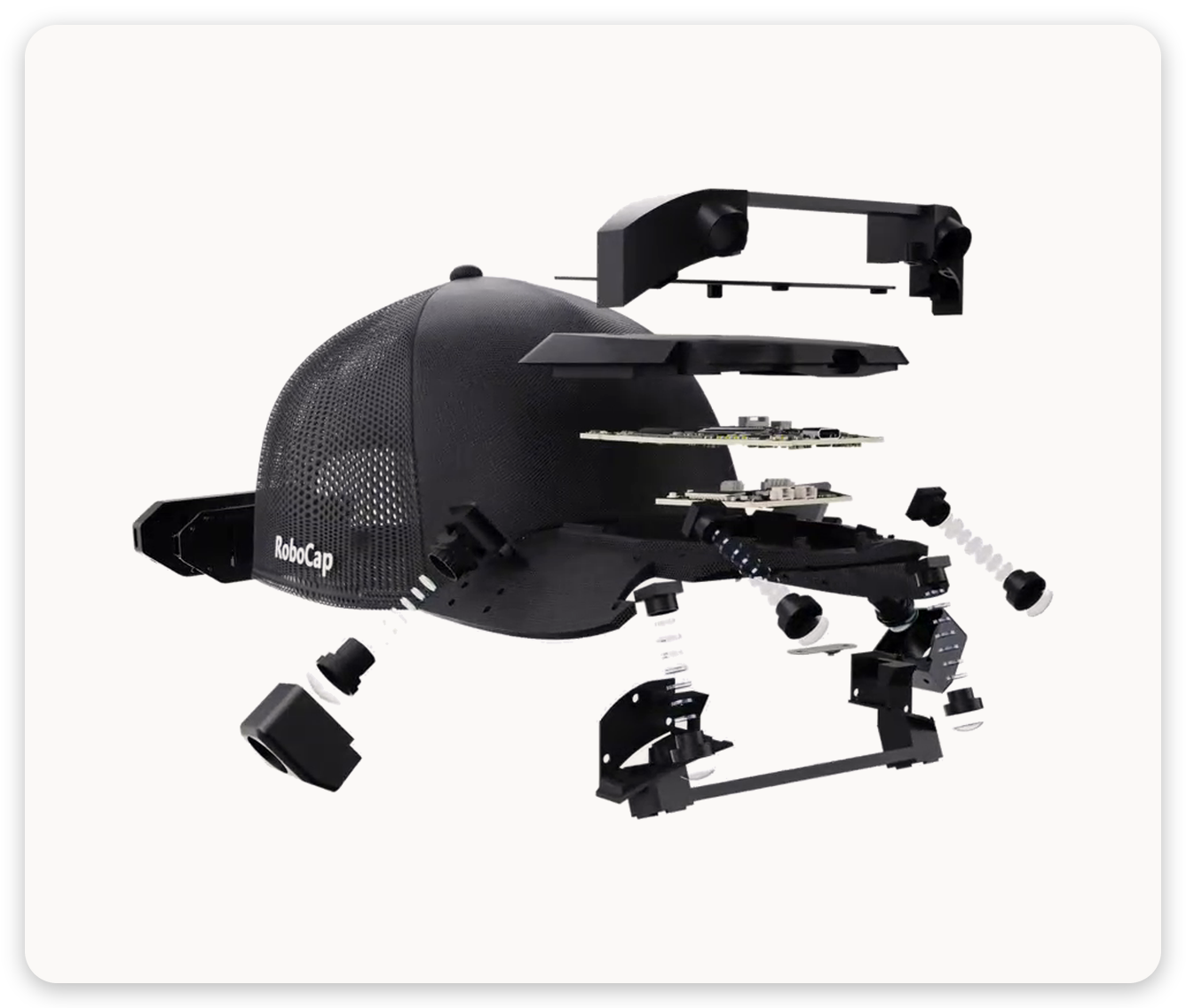}\hfill
    \includegraphics[width=0.32\linewidth]{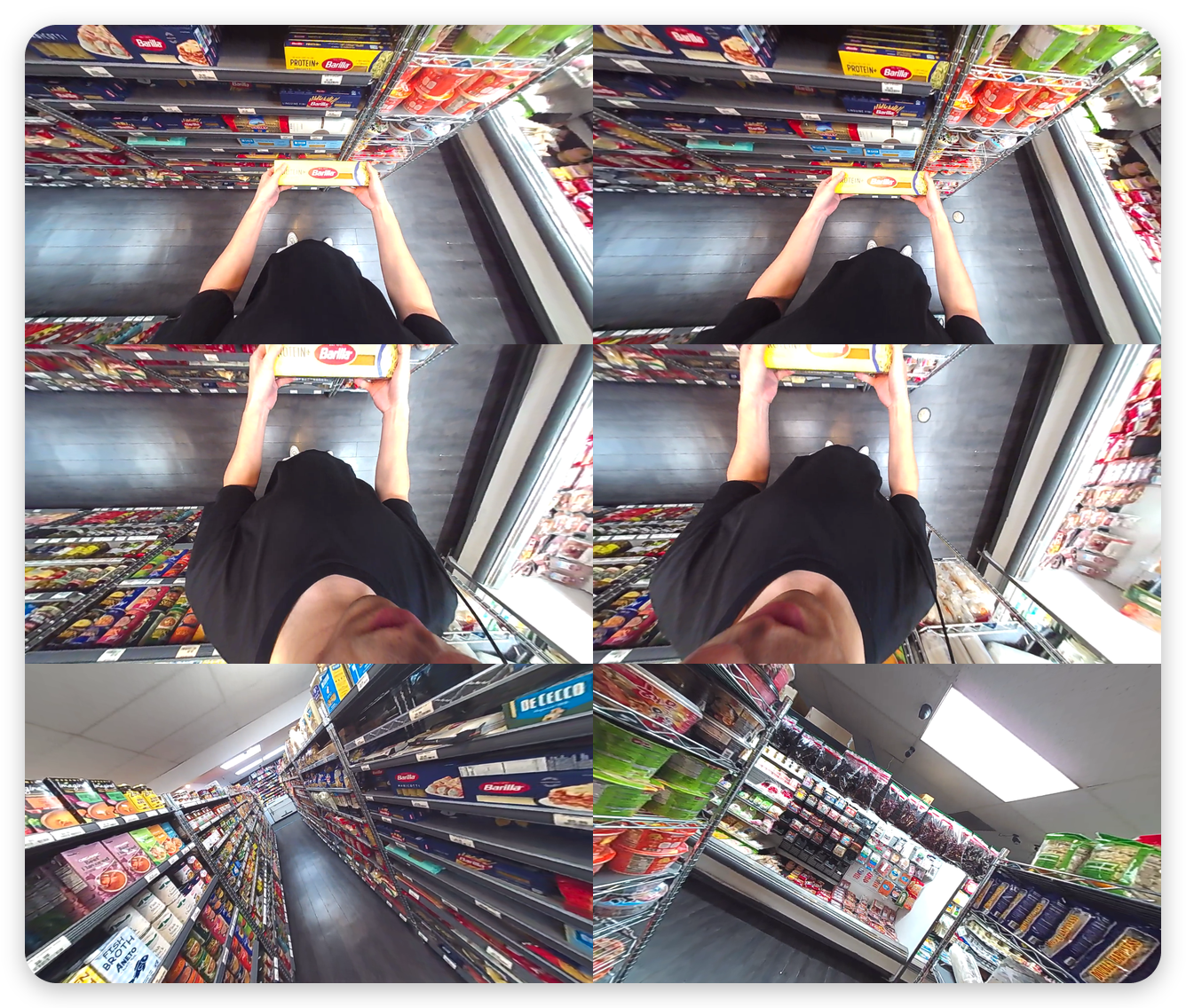}\hfill
    \includegraphics[width=0.32\linewidth]{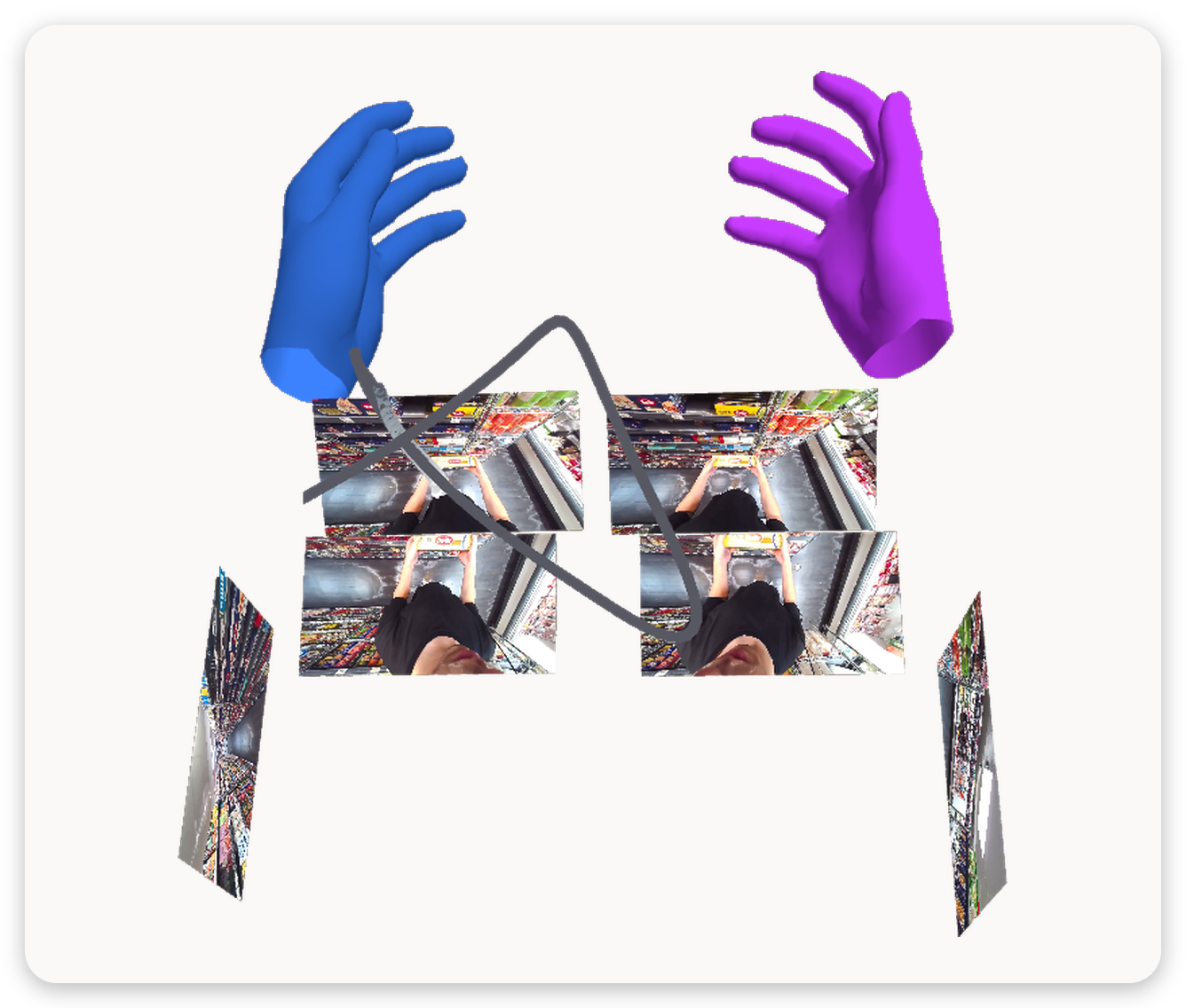}
\end{strip}

\begin{abstract}
Despite its promise for scaling robot learning, egocentric manipulation data is still scarce today. Collection at scale requires vertically integrating ergonomic hardware with centimeter-precise 3D algorithms, at a precision that has not been publicly demonstrated. To address this gap, we introduce RoboCap, a 250\,g six-camera dual-IMU hat designed for in-the-wild egocentric data capture, and the Grounded API, a suite of device-agnostic 3D algorithms tuned for RoboCap. In this report, we demonstrate how hardware, calibration, and 3D algorithms interact to achieve state-of-the-art performance on the public benchmarks: our SLAM across diverse settings and rigs, our depth estimation on egocentric settings, and our hand tracking when adapted to third-party devices.
\end{abstract}

\section{Introduction}
\label{sec:intro}

In the history of large-scale end-to-end robot learning, the field has witnessed the emergence of several data classes, each with known tradeoffs. Simulation scales with compute, but demands human engineering effort that grows with the diversity it aims to produce. Teleoperation captures the exact states and actions a robot requires at inference, at the cost of unnatural and unscalable demonstration through cumbersome robot hardware. UMI deploys only the end-effector for better scalability, but does not scale beyond low-DoF manipulators and requires retargeting. Egocentric data maximizes scalability by equipping humans with an unintrusive headworn device, but recovers states and actions only by estimation and also requires retargeting.

The quality of egocentric data is bounded by the accuracy of its 3D labels, namely the human hand and head poses. Retargeting maps an estimated human pose onto a robot embodiment~\cite{liu2025egozero, kareer2024egomimic, qiu2025humanoidpolicy, hoque2025egodex}, so estimation error enters the policy's training distribution directly as label noise: egocentric capture is a metrology problem before it is a video collection problem, and a large-scale 3D computer vision problem in practice. Centimeter-level accuracy must hold under occlusion, motion blur, varied lighting, and operator diversity, on footage collected outside any lab. The field has not converged on a device the way it has on teleoperation and UMI rigs, and the reason is that the two halves of the problem (form factor and 3D algorithms) have never been available together. Meta's Project Aria glasses~\cite{engel2023aria} are the best form factor candidate, but are gated by application programs, hardware costs, and offline APIs that are difficult to scale. Mixed-reality headsets are heavy, socially intrusive, and expose only opaque on-device tracking whose error characteristics cannot be inspected or improved offline. A team that wants to collect its own egocentric data at the precision policy learning requires cannot currently assemble the stack to do so.

We argue that the estimation problem in egocentric data is an artifact of hardware that was never designed for it. Because consumer smart glasses and mixed-reality headsets optimize for display, audio, and human-facing UX, their sensing suites are byproducts. When the device is instead designed backward from the requirements of the 3D estimation stack--stereo baselines, FoV overlap, hardware-synchronization, global-shutter sensors--the gap between estimated and measured state collapses to a level sufficient for policy learning. As we explain in Section~\ref{sec:3d}, RoboCap's layout is what makes this concrete. Two calibrated stereo pairs supply metric scale that monocular systems can only infer, while wide-angle lateral coverage keeps the hands in frustum through natural motion and keeps the trajectory solver constrained during close-range manipulation.

We also design our 3D algorithms to be device-agnostic. The same binaries run on handheld fisheye rigs, aerial stereo platforms, consumer VR headsets, and research glasses whose divergent cameras share almost no field of view, with no per-dataset tuning. This is not incidental to the argument above but rather evidence for it: portability establishes that our results reflect general 3D estimation rather than models overfit to a single device. Our SLAM, for instance, borrows from both monocular-inertial and stereo-inertial designs, which let it fuse whatever cues a rig provides--six cameras with two calibrated baselines on RoboCap, or two divergent cameras with no stereo overlap on Aria--into a single estimation pipeline.

Our contributions are twofold:
\begin{enumerate}
    \item RoboCap, a purpose-built 250 g six-camera dual-IMU egocentric capture device whose sensing suite is designed minimally and sufficiently for metric 3D estimation, available for purchase.
    \item Grounded API, which serves SLAM, depth estimation, and hand tracking algorithms that achieve state-of-the-art accuracy across diverse visual settings and hardware rigs.
\end{enumerate}

\section{Related work}
\label{sec:related}

The contributions of this paper touch on progress in egocentric wearable hardware and in 3D estimation. In this section we review each in turn.

\subsection{Egocentric wearables}

Egocentric capture hardware spans repurposed consumer devices to purpose-built research platforms. Early large-scale egocentric datasets were collected with head-mounted action cameras~\cite{damen2018epickitchens, grauman2022ego4d}, which provide only monocular RGB with rolling shutter and no inertial or calibration guarantees, precluding metric 3D annotation. Mixed-reality headsets such as Apple's Vision Pro and Meta's Quest 3 offer richer sensing and supply head and hand pose estimation~\cite{hoque2025egodex, banerjee2024hot3d}, but these devices are heavy, socially intrusive, and expose their sensing only through on-device tracking APIs.

The closest platform in spirit to ours is Meta's Project Aria~\cite{engel2023aria}: lightweight research glasses with calibrated RGB, monochrome SLAM cameras, and IMUs, that have powered a family of egocentric datasets~\cite{lv2024ariaeveryday, ma2024nymeria, pan2023adt, banerjee2024hot3d}. Aria validates the thesis that calibrated multi-camera visual-inertial wearables enable high-quality 3D annotation, and prior works in robotics have demonstrated its use in learning robot manipulation policies from human data~\cite{liu2025egozero, kareer2024egomimic}. Aria remains, however, a general-purpose research device rather than a scalable manipulation capture platform: its cameras are laid out for scene understanding rather than for the near field, and its complexity has kept it relatively inaccessible for robot learning at scale.

There is also a lineage of research prototypes aimed at estimating human poses. For example, ~\cite{rhodin2016egocap, xu2019mo2cap2, akada2022unrealego} propose head- or cap-mounted downward-facing fisheye rigs, but for target body pose rather than manipulation. More recent works combine mocap gloves with an egocentric device to estimate hand pose, but are too clunky and brittle to occlusions~\cite{wang2024dexcap, tao2026dexwilddexteroushumaninteractions}. These prototypes introduce intrusive hardware in an attempt to measure hand pose accurately, but thereby sacrifice the naturalness and scalability. Inevitably, a single egocentric device must natively compute accurate hand poses directly from vision in order to maximize the scalability of egocentric data capture.

\begin{figure*}[t]
  \centering
  \includegraphics[width=\textwidth]{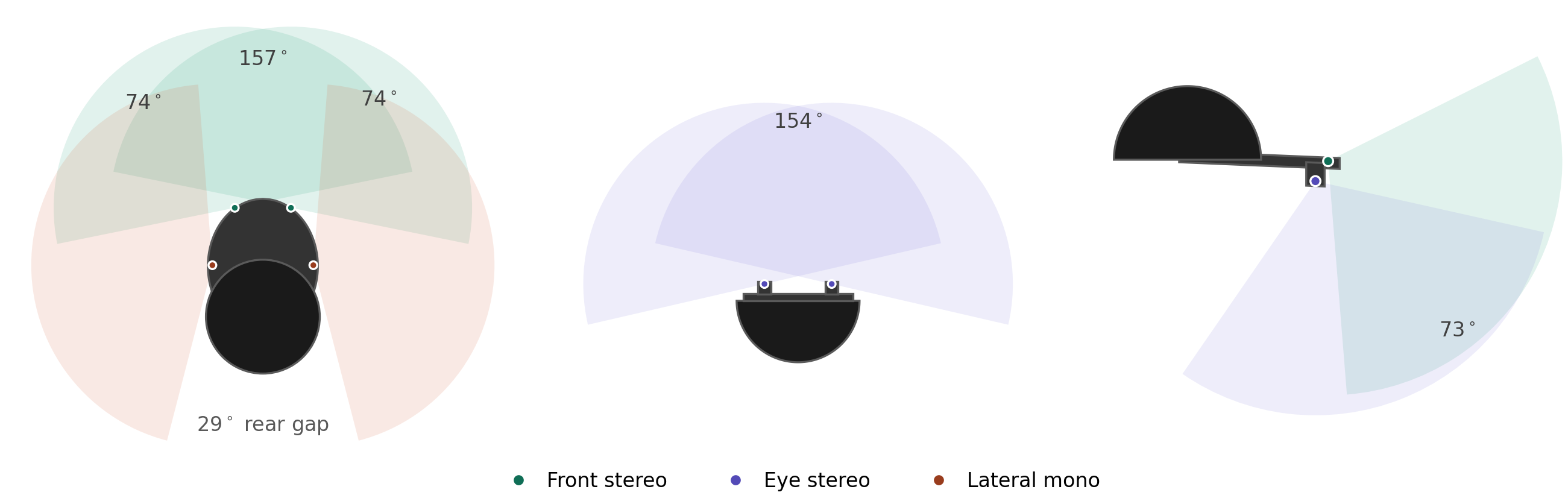}
  \caption{RoboCap camera coverage. Left: azimuth coverage of the front stereo and lateral cameras. Center: azimuth coverage of the eye stereo cameras, visualized forward for legibility. Right: elevation coverage of the front and eye pairs. FoV-overlap in degrees at infinity.}
  \label{fig:sensor_geometry}
\end{figure*}

\subsection{3D estimation}

\paragraph{Camera calibration.}
Multi-sensor rigs are typically calibrated offline by continuous-time batch optimization over a fiducial target~\cite{furgale2013kalibr}, jointly recovering intrinsics, camera--IMU extrinsics, and time offsets. Factory calibration degrades in the field through mechanical and thermal stress, motivating online self-calibration from natural correspondences during operation~\cite{dang2009continuous, lowe2004sift}. As we demonstrate in Section~\ref{sec:online_calibration}, this drift becomes important when needing to achieve sub-centimeter 3D estimation errors.

\paragraph{Visual-inertial SLAM.}
Classical sparse visual-inertial systems establish the standard factor-graph formulation fusing feature reprojection with IMU preintegration~\cite{campos2021orbslam3, qin2018vinsmono, leutenegger2015okvis, usenko2020basalt, rosinol2020kimera, geneva2019openvins}, with global consistency recovered through appearance-based place recognition~\cite{galvez2012dbow2} and pose-graph optimization~\cite{grisetti2010tutorial}. GPU-accelerated implementations extend this to multi-camera rigs at real-time rates~\cite{korovko2025cuvslam}. A learned line trades classical sparse optimization for dense correspondence and feed-forward geometry~\cite{teed2021droidslam, murai2025mast3rslam, wang2025vggt}, while rendering-based backends push offline reconstruction quality at significant compute cost~\cite{zhu2026vigsslamvisualinertialgaussian, cao2026mcgsslammulticameraslamframework}.

These systems are evaluated on a small set of standard benchmarks with external ground truth, spanning aerial stereo-inertial, handheld fisheye, multi-sensor rigs~\cite{burri2016euroc, schubert2018tumvi, gao2022vector}, and most recently consumer VR headsets~\cite{demayo2025monado}, the closest existing benchmark to the egocentric regime. Egocentric head motion is a distinctly adversarial setting for all of them: rapid rotation, close-range parallax, dynamic scene content from the wearer's own hands, and irregular motion during initialization. This motivates both our multi-camera wide-FoV configuration and our offline post-processing of the estimated trajectory.

\paragraph{Stereo depth estimation.}
Deep stereo matching progressed from cost-volume CNNs to iterative recurrent refinement~\cite{chang2018psmnet, lipson2021raftstereo, li2022crestereo, xu2023igev, wang2024selectivestereo}, but these models require per-domain fine-tuning and degrade sharply out of distribution. A more recent class trains on large-scale high-fidelity synthetic corpora, bootstrapping monocular foundation priors to achieve strong zero-shot generalization for the first time~\cite{wen2025foundationstereo, wen2026fastfoundationstereo}. Monocular metric alternatives~\cite{bhat2023zoedepth, yin2023metric3d, bochkovskii2024depthpro, yang2024depthanything, yang2024depthanythingv2} remove the calibrated-baseline requirement, but their metric scale is inferred from learned priors rather than measured from geometry, which is exactly the failure mode our hardware exists to eliminate.

\begin{table*}[t]
  \centering
  \small
  \setlength{\tabcolsep}{4pt}
  \begin{tabular}[t]{@{}ll@{}}
    \toprule
    \multicolumn{2}{@{}l}{\textbf{Cameras} (6, hardware-synchronized)} \\
    \midrule
    Sensor & Global-shutter RGB, 10-bit, 3.0\,\textmu m px \\
    Frame resolution & $1920{\times}1080$ \\
    Frame rate & 30\,Hz \\
    Lens & Kannala-Brandt 4 \\
    Focal length & $f = 590$--$630$\,px \\
    Exposure & Per-camera AE, bounded $\leq 10$\,ms \\
    Stereo optics & Fisheye, $180^\circ{\times}112^\circ$ ($216^\circ$ diag.) \\
    Lateral optics & Wide, $161^\circ{\times}98^\circ$ ($179^\circ$ diag.) \\
    Front stereo & $2\times$ 86.8\,mm baseline, $157^\circ$ overlap \\
    Eye stereo & $2\times$ 106.2\,mm baseline, $154^\circ$ overlap \\
    Lateral & $2\times$ mono, $85^\circ$ off forward axis, $30^\circ$ down \\
    \addlinespace
    \midrule
    \multicolumn{2}{@{}l}{\textbf{Magnetometer} (1, crown center)} \\
    \midrule
    Rate & 3-axis, 100\,Hz, time-aligned (auxiliary) \\
    \bottomrule
  \end{tabular}
  \hfill
  \begin{tabular}[t]{@{}ll@{}}
    \toprule
    \multicolumn{2}{@{}l}{\textbf{IMUs} (2, opposite sides of crown, 6-axis)} \\
    \midrule
    Rate & 200\,Hz, hardware-synchronized interrupt sampling \\
    Gyro.\ noise density & $6{\times}10^{-4}\,\mathrm{rad\,s^{-1}/\sqrt{Hz}}$ \\
    Gyro.\ bias walk & $3{\times}10^{-5}\,\mathrm{rad\,s^{-2}/\sqrt{Hz}}$ \\
    Accel.\ noise density & $7{\times}10^{-3}\,\mathrm{m\,s^{-2}/\sqrt{Hz}}$ \\
    Accel.\ bias walk & $2{\times}10^{-4}\,\mathrm{m\,s^{-3}/\sqrt{Hz}}$ \\
    Camera--IMU offset & Residual $\leq 3$\,ms \\
    \addlinespace
    \midrule
    \multicolumn{2}{@{}l}{\textbf{System}} \\
    \midrule
    Reference clock & 24\,MHz TCXO, $\pm 1.81$\,ppm \\
    Trigger & Single PWM, 1-to-6 fan-out to all cameras \\
    Power & 4--6\,W \\
    Weight & 250\,g \\
    Runtime & 6\,h per 10{,}000\,mAh \\
    Storage & 214.5\,GB onboard \\
    \bottomrule
  \end{tabular}
  \caption{RoboCap sensor suite. The two stereo pairs use fisheye optics, the lateral cameras a wide lens. Stereo overlap is the binocular field at infinity from the calibrated extrinsics. IMU noise parameters are fleet-level estimates from per-device continuous-time calibration~\cite{furgale2013kalibr}.}
  \label{tab:sensors}
\end{table*}

\paragraph{Hand tracking.}
Parametric hand recovery is dominated by the MANO model~\cite{romero2017embodied}. Monocular mesh recovery has scaled steadily from early regression and real-time keypoint systems~\cite{rong2021frankmocap, zhang2020mediapipehands} to transformer-based reconstructors trained at scale~\cite{pavlakos2024hamer, potamias2024wilor}. Critically, all monocular methods regress pose under a weak-perspective camera, which infers shape and depth from apparent hand size against a learned prior. Not only are monocular outputs metrically ambiguous, they are also not consistent enough across views for accurate triangulation, which inevitably must be learned with neural network design~\cite{yang2023poem, iskakov2019learnable}. World-space extensions recover trajectories from moving egocentric video, but are still susceptible to scale ambiguity as well as SLAM inaccuracies~\cite{zhang2025hawor, yu2025dynhamr}.

Multi-view systems resolve the metric ambiguity of monocular prediction, but at the cost of added hardware dependency. Production-grade egocentric trackers fuse several head-mounted fisheye views into a single metric prediction~\cite{han2020megatrack, han2022umetrack}, and the multi-view egocentric benchmarks built on their data~\cite{banerjee2024hot3d, rim2026show3dcapturingscenes3d} evaluate this setting directly. These systems derive their training labels from motion-capture installations or closed on-device trackers, so annotation cost scales with instrumentation rather than with compute. Furthermore, motion capture confines collection to instrumented spaces, which recent work~\cite{rim2026show3dcapturingscenes3d} addresses by building a large exoskeleton cage. Yet the difficulty of hand tracking is in-the-wild, where metric accuracy must persist through occlusion, motion blur, and operator diversity. Our method solves this by enforcing geometric consistency from monocular priors, allowing us to generate a large corpus of in-the-wild pre-training data without any motion capture. Section~\ref{sec:hand_tracking} evaluates models trained on this corpus against baselines trained on motion-capture-derived data~\cite{han2022umetrack, banerjee2024hot3d, rim2026show3dcapturingscenes3d}.

\section{Device}
\label{sec:device}

The sensing suite is specified by what metric 3D estimation requires at manipulation range. Depth must be measured rather than inferred, which requires at least one calibrated stereo pair with a baseline wide enough that disparity resolves centimeters at arm's length. The hands must remain in frustum through natural motion, including reaches that leave the forward field, which requires wide-angle coverage well beyond the forward-facing cone. Multi-view fusion is only valid if views are simultaneous, which requires hardware-triggered global-shutter sensors rather than per-channel scheduling of rolling-shutter streams. And metric scale is only as good as the calibration it rests on, which requires per-unit factory characterization rather than a nominal model shared across a fleet. Table~\ref{tab:sensors} summarizes the resulting suite: six hardware-synchronized global-shutter cameras in two calibrated stereo pairs with two lateral wide-angle views, and two IMUs at opposite sides of the crown.

The two requirements pull in different directions, so the cameras do not share optics. The stereo pairs carry fisheye lenses spanning $180^\circ{\times}112^\circ$ ($216^\circ$ diagonal), which keeps the hands within a metric stereo frustum across the full reach envelope; the front pair sits on a 86.8\,mm baseline with $157^\circ$ of binocular overlap and the eye pair on a 106.2\,mm baseline with $154^\circ$. The lateral cameras instead carry a wide lens of $161^\circ{\times}98^\circ$ ($179^\circ$ diagonal), mounted $85^\circ$ off the forward axis and pitched $30^\circ$ down, where the priority is catching reaches that leave the stereo frustum entirely and keeping the trajectory solver constrained during close-range manipulation rather than triangulating them. Each lateral view retains roughly $80^\circ$ of overlap with the common stereo field, leaving a $20^\circ$ gap directly behind the wearer.

The hat form factor was chosen against glasses, headsets, and phone rigs to accommodate that suite. A hat distributes weight over the crown, permitting a larger sensor and battery payload than eyewear can carry without discomfort; it requires no display, audio, or interaction hardware, so cost is dominated by cameras and IMUs; and it is socially unobtrusive enough for collection outside a lab, where the data this device exists to capture is produced. The result is 250\,g, worn for hours without adjustment.

The device draws 4--6\,W in typical operation and is powered by a separate tethered battery bank; a 10{,}000\,mAh bank sustains roughly six hours of continuous recording. A small onboard lithium-ion battery serves as a fail-safe. If the tether is severed mid-session, it keeps the device alive to finalize and save the recording, and it powers data offload when no bank is attached, so no session is lost to a power interruption. Sessions are encoded onboard (H.265, YUV 4:2:0, 4\,Mbps per stream) to 214.5\,GB of internal storage (10+ hours of recording) and offloaded via USB~3 at 60--200\,MB/s.

\begin{figure*}[t]
    \centering
    \includegraphics[width=0.30\linewidth]{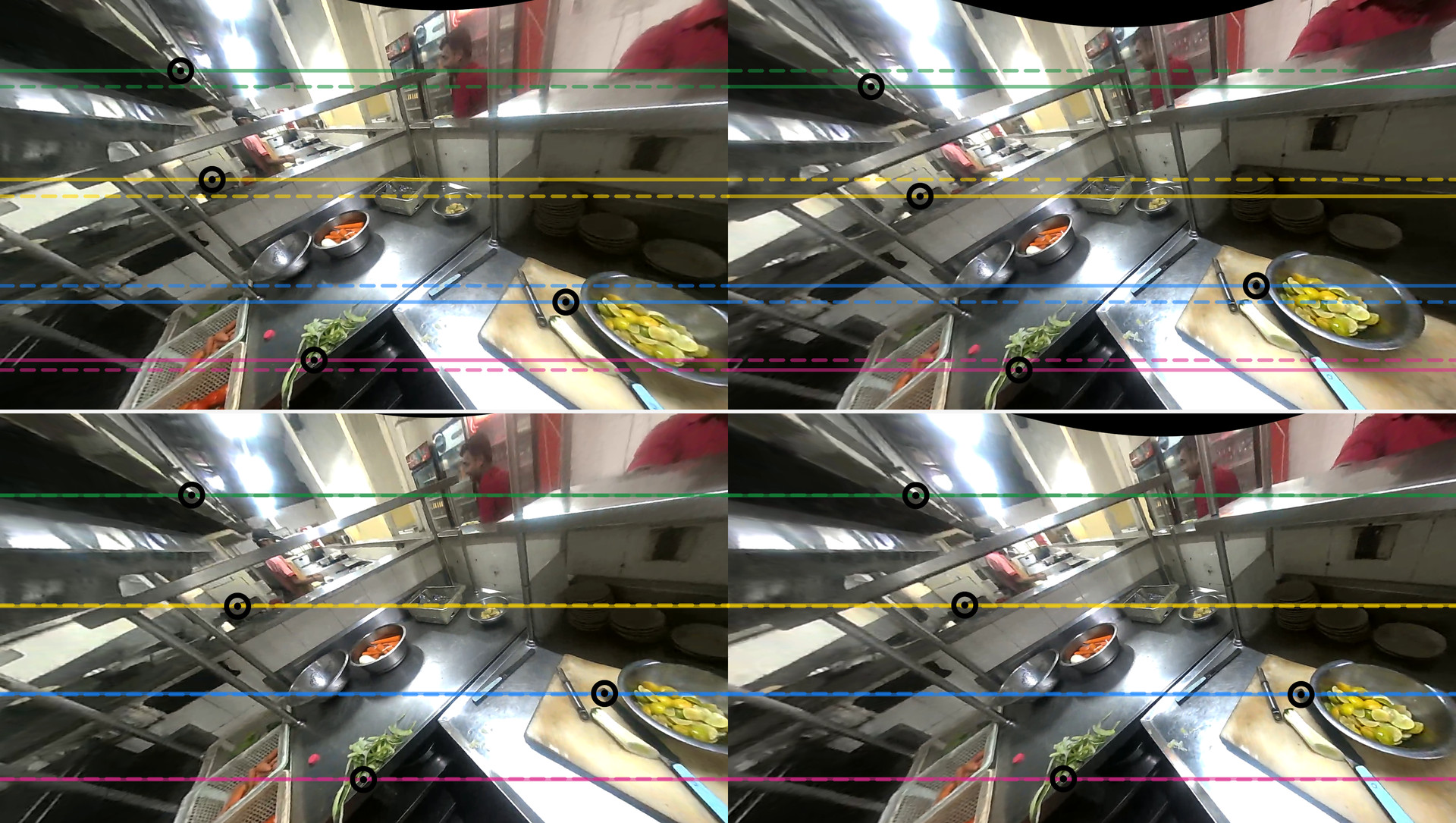}\hfill
    \includegraphics[width=0.30\linewidth]{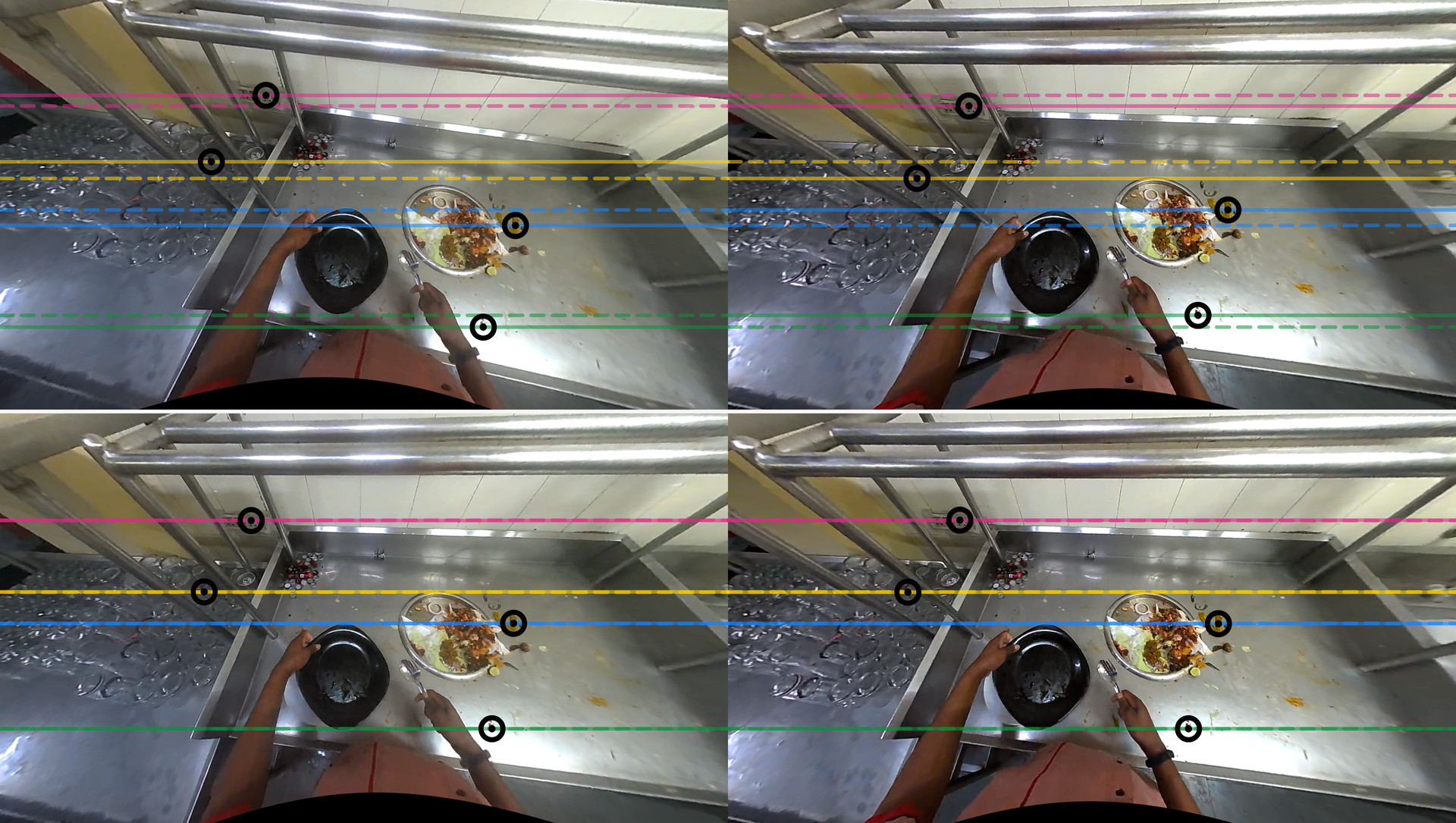}\hfill
    \includegraphics[width=0.30\linewidth]{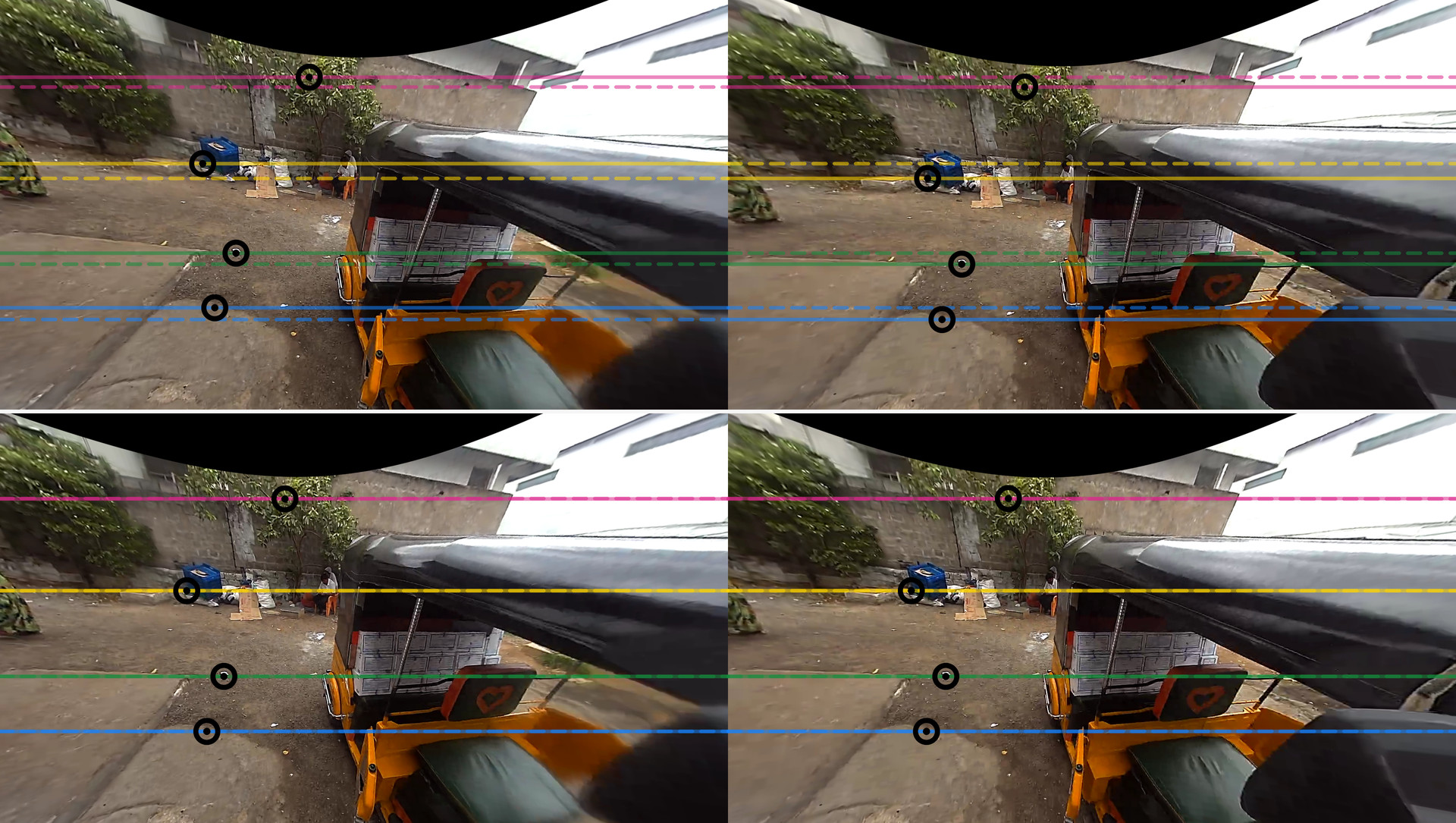}
    \caption{Epipolar residuals before (top) and after (bottom) self-calibration. Identical features are marked in the left and right cameras with a black dot, and epipolar lines are drawn across each pair. Self-calibration eliminates almost all epipolar drift and improves 3D estimation.}
    \label{fig:self_calibration_epi}
\end{figure*}

\begin{figure*}[t]
    \centering
    \includegraphics[width=0.30\linewidth]{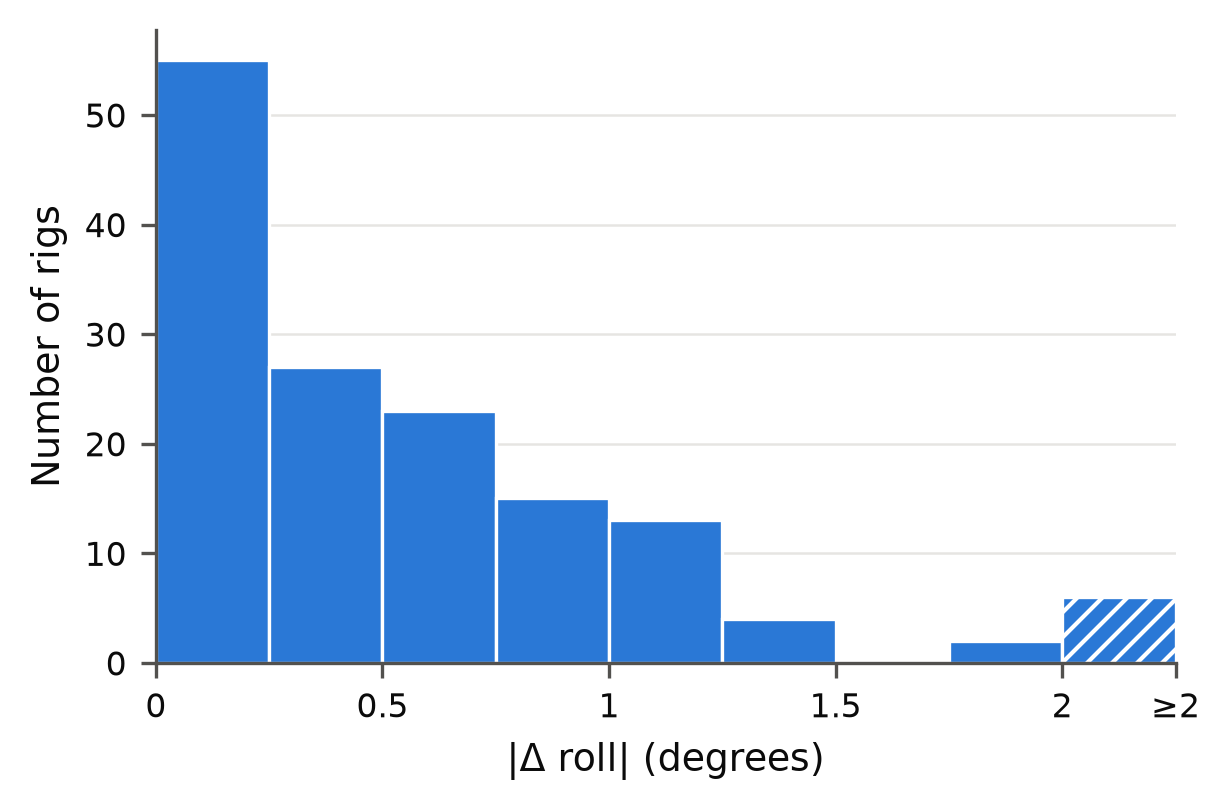}\hfill
    \includegraphics[width=0.30\linewidth]{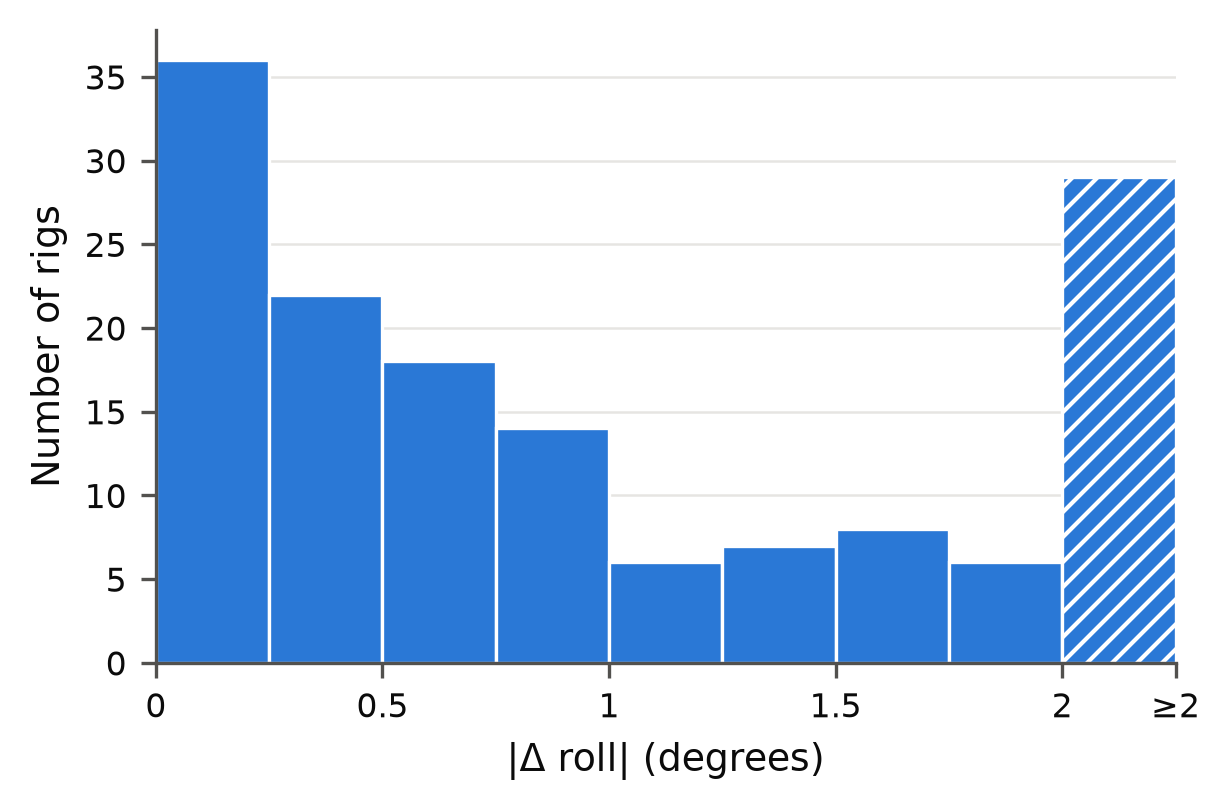}\hfill
    \includegraphics[width=0.30\linewidth]{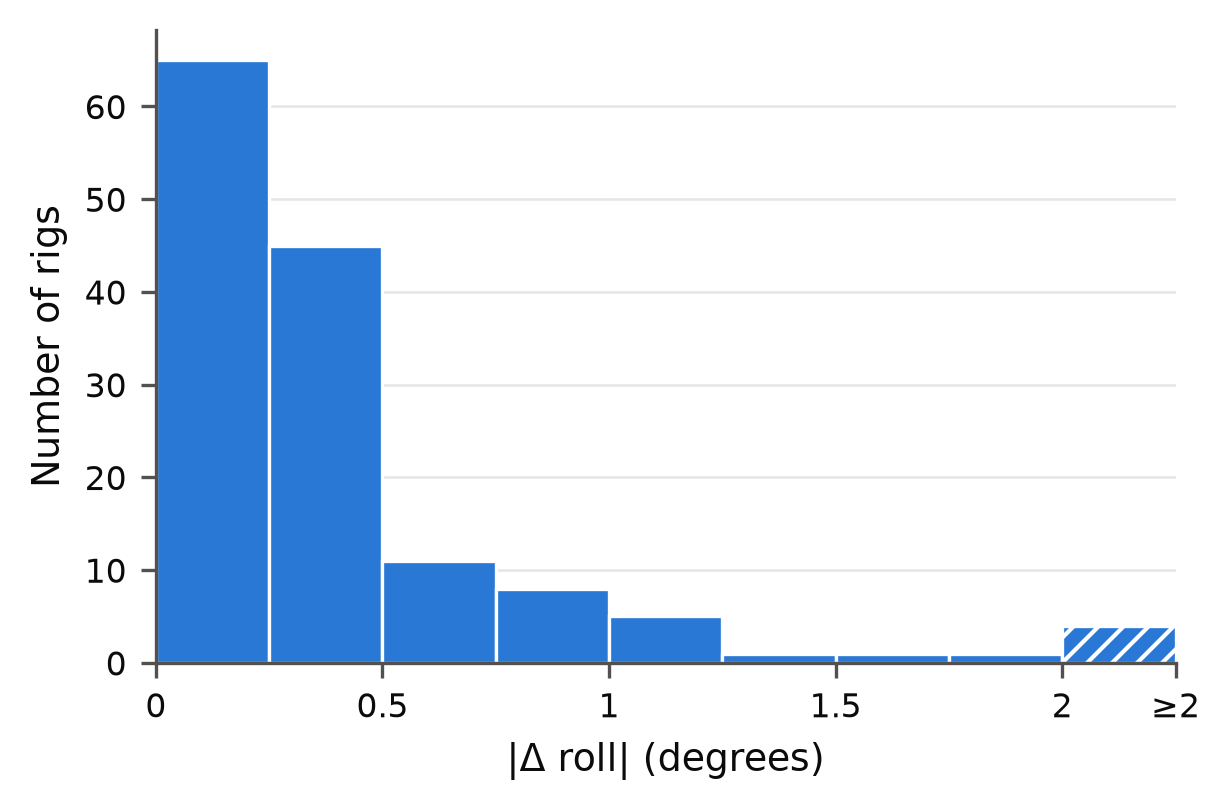}
    \caption{Histograms of the epipolar drift in the eye pair (left), front pair (middle), and front-eye pairs (right), over a subset of 150 deployed devices. We find that rotational correction (Section~\ref{sec:online_calibration}) is sufficient to restore epipolar drift, which follows a Gaussian centered around 0$^\circ$.}
    \label{fig:self_calibration_hist}
\end{figure*}

\subsection{Calibration}
\label{sec:factory_calibration}

Every unit is individually calibrated at the factory before shipment. An operator sweeps the device through a calibration motion sequence against a stationary AprilGrid target ($6{\times}6$, 55\,mm tags) simultaneously covering all camera groups and both IMUs. IMU noise models are identified by Allan-variance analysis; camera intrinsics (pinhole + Kannala--Brandt KB-4 distortion, the standard choice at these field-of-view angles), camera--IMU extrinsics, and time offsets are solved by continuous-time batch optimization~\cite{furgale2013kalibr}. Because the two lens types are calibrated against the same target sweep, the recovered focal lengths span $590$--$630$\,px across the fleet rather than a single nominal value. Units are accepted only within per-parameter tolerance gates for each of camera extrinsics and intrinsics reprojection, gyroscope, and accelerometer--and units failing any gate are reworked rather than shipped.


Factory calibration is performed at room temperature and the device applies no temperature-dependent parameter correction, so calibration degrades in the field through mechanical and thermal stress. At manipulation range this drift is not a second-order effect: a pixel of disparity error translates to roughly a centimeter of depth at $0.65$\,m. Rather than tighten the mechanical design against a drift that cannot be eliminated, we correct it per session from natural correspondences, which Section~\ref{sec:online_calibration} describes and quantifies.

\subsection{Time synchronization}
\label{sec:sync}

All six cameras are exposure-triggered from a single PWM frame-synchronization signal generated by the main controller and fanned out through a 1-to-6 clock buffer, so simultaneity holds by construction rather than by per-channel scheduling; the measured camera--camera offset is 0\,ms at timestamping granularity. Frames are timestamped at exposure completion on the device clock. The IMUs stream autonomously at 200\,Hz under hardware-synchronized interrupt sampling and are timestamped on arrival in the same clock domain, with the camera--IMU offset identified per device by continuous-time calibration~\cite{furgale2013kalibr} to a residual of $\leq 3$\,ms, verified from natural recordings by correlating gyroscope rates against image-derived rotation.

The device clock is disciplined by a 24\,MHz TCXO with a $\pm 1.81$\,ppm worst-case budget, bounding absolute timeline drift to ${\approx}\pm 6.5$\,ms per hour. Because all sensors share this clock, that drift affects the absolute timeline only and not inter-sensor synchronization, which is what multi-view fusion and visual-inertial estimation depend on.

\begin{figure*}[t]
  \centering
  \includegraphics[width=0.9\textwidth]{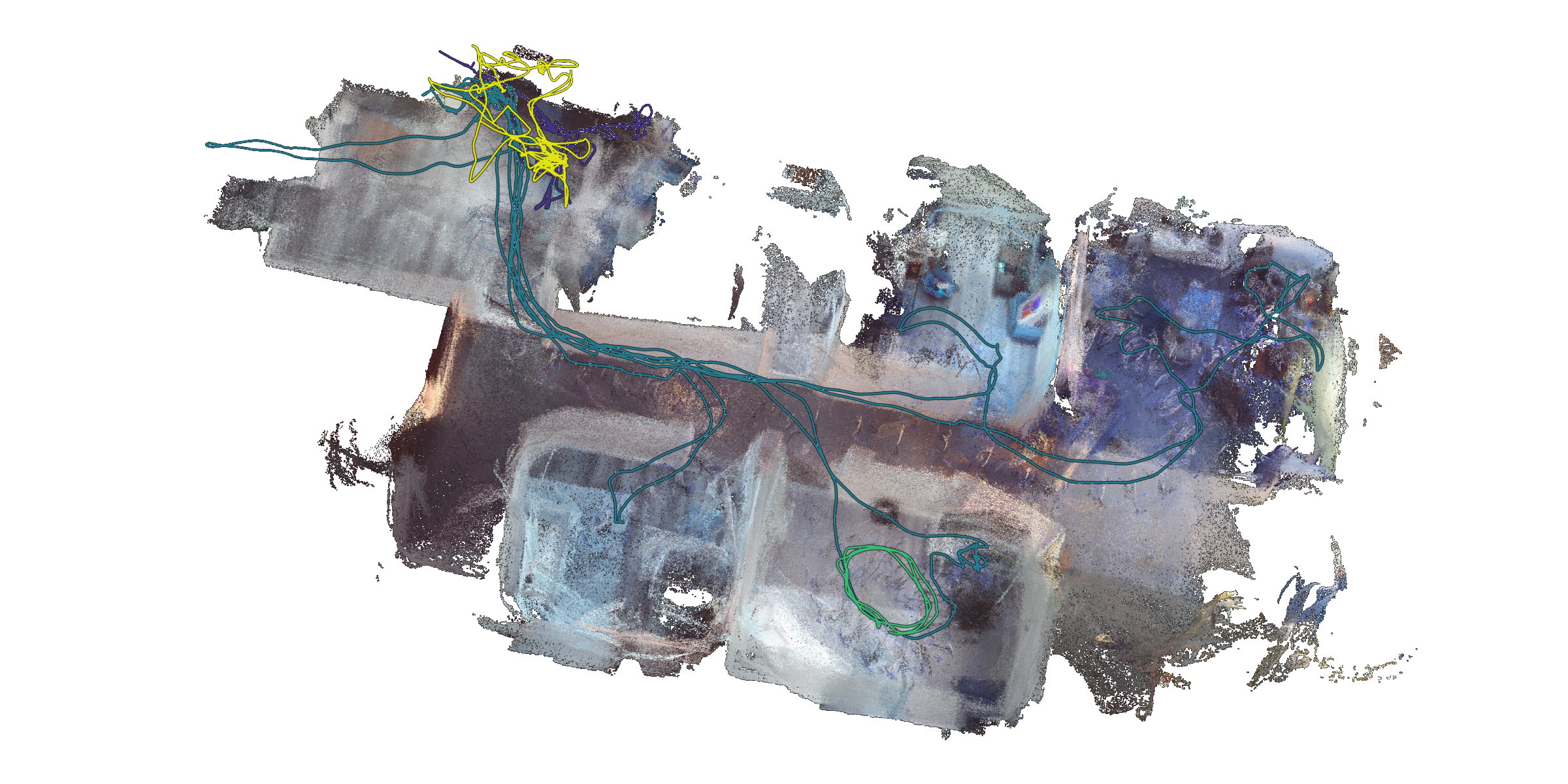}
  \caption{Scene reconstruction over 4 RoboCap sessions. For each session, we generate point clouds (Section~\ref{sec:depth}) and project them into the SLAM world frame, leveled with the estimated gravity direction (Section~\ref{sec:slam}). Then, we solve multi-session frame alignment with RANSAC on 3D-projected SIFT features.}
  \label{fig:scene_reconstruction}
\end{figure*}

\begin{table*}[t]
    \centering
    \small
    \begin{tabular}{llcccccc}
    \toprule
    Method & Inputs & Short & Medium & Long & Low light & Moving & Avg. \\
    \midrule
    \textcolor{gray}{Aria's SLAM (r)$^{\ddagger}$} & \textcolor{gray}{bino, IMU} & \textcolor{gray}{90.7} & \textcolor{gray}{78.5} & \textcolor{gray}{70.9} & \textcolor{gray}{84.2} & \textcolor{gray}{55.0} & \textcolor{gray}{75.9} \\
    \midrule
    ORB-SLAM3 (r)~\cite{campos2021orbslam3} & mono, IMU & 23.0 & 10.9 & 11.2 & 3.1 & 2.0 & 10.0 \\
    OKVIS2 (r)~\cite{leutenegger2022okvis2} & bino, IMU & 20.0 & 11.6 & 2.6 & 14.5 & 4.7 & 10.7 \\
    OpenVINS+Maplab (r) & bino, IMU & 27.7 & 23.4 & 12.8 & 19.8 & 13.9 & 19.5 \\
    Best of other submissions & mono, IMU & 34.2 & 29.3 & 16.8 & 25.4 & 15.5 & 24.2 \\
    \midrule
    Ours & bino, IMU & \textbf{80.2} & \textbf{61.6} & \textbf{59.9} & \textbf{67.7} & \textbf{41.9} & \textbf{62.3} \\
    \quad pose recall @5\,m (\%) & & 99.9 & 96.2 & 99.3 & 99.9 & -- & \\
    \bottomrule
    \end{tabular}
    \caption{Official LaMAria test-set results (control-point score, higher is better). (r): reproduced by the benchmark authors. ``Best of other submissions'' is the highest score per category over all other public entries. $^{\ddagger}$Shown in gray as a reference and excluded from the ranking: Aria's SLAM is the device manufacturer's own system, developed together with the glasses, and is the solution used to initialize the benchmark's ground-truth pipeline (its pose recall is therefore not reported). Bold marks the best result among the remaining methods.}
    \label{tab:slam_lamaria}
\end{table*}

\section{3D estimation}
\label{sec:3d}

The most common way to retarget motions from human data is to define a unified action space across embodiments as the 3D end-effector pose. Therefore, the role of the 3D stack is to recover this pose, together with the geometry needed to contextualize it, at metric scale in a consistent world frame. In this section, we describe our SLAM, depth estimation, and hand tracking implementations that we have evaluated to be state-of-the-art on several cross-rig and in-rig benchmarks.

\subsection{Online self-calibration}
\label{sec:online_calibration}

Centimeter precision demands tight alignment between the matcher and the device calibration. One pixel of disparity corresponds to ${\sim}1$\,cm of depth at $0.65$\,m, and stereo error grows as $z^2/(fb)$. In practice, the rig drifts slightly from its factory calibration through mechanical and thermal stress. Across our fleet, the relative rotation shifts by $0.2$--$1.1^\circ$ (median ${\approx}0.5^\circ$), producing $1$--$5$\,px of epipolar residual and a distance-dependent depth bias of $1$--$2$\,cm at manipulation range. Our observations are consistent with what Aria reports ${\approx}25$\,arcmin of instantaneous deformation between its stereo cameras from the act of removing the device~\cite{engel2023aria}.

Drift is dominated by rotation, so we only estimate the residual rotations once per recording session from natural correspondences~\cite{lowe2004sift, dang2009continuous} with no calibration target and no operator intervention. The left-front camera is the anchor and is never moved. Each pair is constrained by its own correspondences and fit independently against the Sampson epipolar distance rather than the vertical residual of a rectified pair. Corrections are accepted only if they improve a held-out residual and fall within ${\approx}1^\circ$ of that device's own factory value. Figure~\ref{fig:self_calibration_hist} shows the spread of online self-calibrated rotations and Figure~\ref{fig:self_calibration_epi} visualizes epipolar lines before and after online calibration.

\begin{figure*}[!t]
\centering
\includegraphics[width=\linewidth]{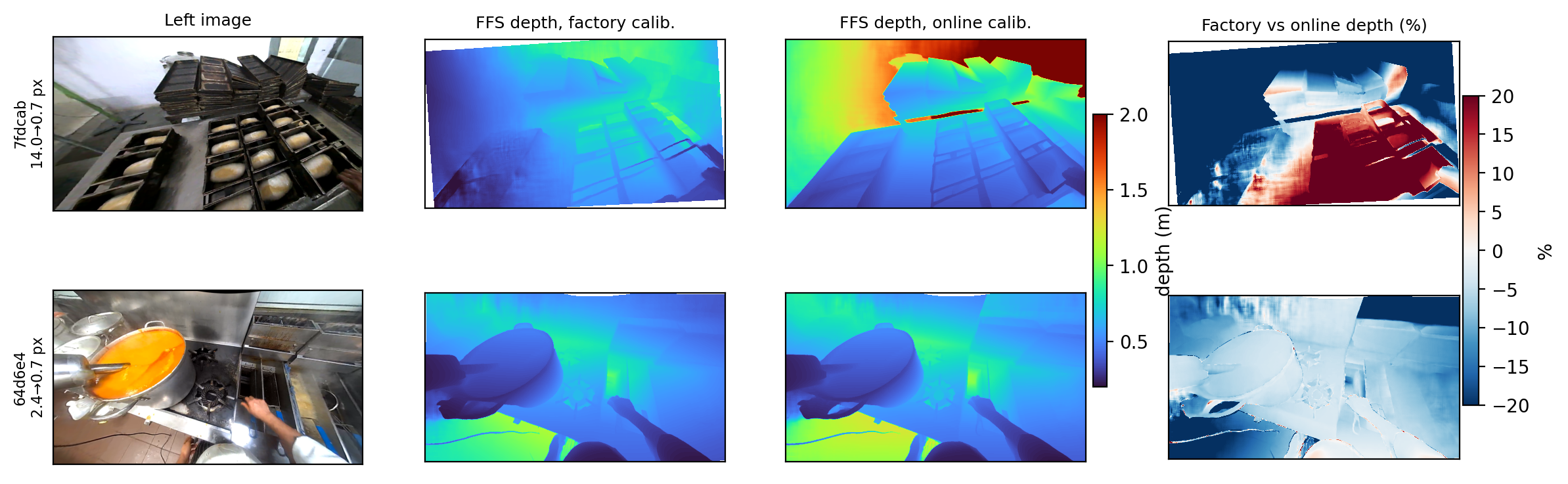}
\captionof{figure}{FFS depth with factory vs.\ online calibration, and the depth change between them. Factory calibration does not visibly degrade FFS depth maps: they remain sharp and complete, but are systematically distorted in scale and shape. Online calibration leaves the maps' appearance unchanged while removing this distortion, halving depth error within reach. See Figure~\ref{fig:depth_calib2} for more examples.}
\label{fig:depth_calib}

\vspace{12pt}

\small
\setlength{\tabcolsep}{4pt}
\begin{minipage}[t]{0.56\textwidth}
\centering
\begin{tabular}{lrrrrr}
\toprule
 & & \multicolumn{2}{c}{FFS~\cite{wen2026fastfoundationstereo}} & \multicolumn{2}{c}{FS~\cite{wen2025foundationstereo}} \\
\cmidrule(lr){3-4}\cmidrule(lr){5-6}
Pixels & \% px & AbsRel$\downarrow$ & $\delta_{1.25}\uparrow$ & AbsRel$\downarrow$ & $\delta_{1.25}\uparrow$ \\
\midrule
All & 100.0 & 0.455 & 69.7 & 0.278 & 75.5 \\
Visible in both views & 11.8 & 0.016 & 99.6 & 0.015 & 99.7 \\
Visible in one view & 26.8 & 0.297 & 61.2 & 0.171 & 77.8 \\
\bottomrule
\end{tabular}
\captionof{table}{Zero-shot depth on UnrealEgo2~\cite{akada20243dhumanposeperception}. In both views: pixels at 0.3--1.0\,m with a correspondence in the right image. In one view: pixels without correspondence. FS and FFS are comparable where there is correspondence.}
\label{tab:ue2_models}
\end{minipage}\hfill
\begin{minipage}[t]{0.40\textwidth}
\centering
\begin{tabular}{lrr}
\toprule
 & \multicolumn{2}{c}{Calibration} \\
\cmidrule(lr){2-3}
Distance & Factory (Sec.~\ref{sec:factory_calibration}) & Online (Sec.~\ref{sec:online_calibration}) \\
\midrule
$<$0.8\,m & 0.023 & 0.011 \\
$\geq$0.8\,m & 0.029 & 0.016 \\
All & 0.025 & 0.012 \\
\bottomrule
\end{tabular}
\captionof{table}{Depth AbsRel of zero-shot FFS on a ChArUco board seen by the RoboCap front pair (1{,}277 frames, 106{,}602 corners, same frames for both calibrations).}
\label{tab:charuco}
\end{minipage}
\end{figure*}

\subsection{SLAM}
\label{sec:slam}

Head motion is estimated by a multi-camera visual-inertial SLAM system of our own, built from established components: a GPU-accelerated patch-based optical-flow front-end and square-root sliding-window estimator~\cite{usenko2020basalt}, appearance-based place recognition~\cite{korovko2025cuvslam}, and an offline keyframe bundle adjustment~\cite{leutenegger2022okvis2}. One configuration is used for every device and dataset; the few rig-dependent choices are derived from the rig itself.

All cameras feed one estimator. Cameras with overlapping fields of view form a stereo pair, giving landmarks metric depth from their first frame; on RoboCap this is the front pair, while the lateral cameras track their own landmarks monocularly and keep the estimator constrained when the hands cover the front pair. Because the patch tracker was designed for VGA-class images, images are downsampled to at most 1000\,px wide; at full resolution, tracks die under motion blur and low light and the estimator diverges in unlit rooms. The estimator is initialized from motion and re-initialized when a divergence check detects implausible biases or pose jumps~\cite{qin2018vinsmono}. We apply online calibration per Section~\ref{sec:online_calibration}.

Reported trajectories come from an offline visual-inertial bundle adjustment over all keyframes, run once all revisits are known, in which recognized revisits enter as merged landmarks and relative-pose constraints. The causal estimate produced during capture uses the same front-end and estimator, with pose-graph loop closure in place of this refinement.

\paragraph{Evaluation on LaMAria.}
LaMAria~\cite{krishnan2025benchmarkingegocentricvisualinertialslam} is the closest public match to our setting: a head-worn multi-camera visual-inertial device recording everyday walking, at city scale. Its 63 test recordings total 27.5\,h and 76\,km, with single walks of up to 48 minutes and 2.2\,km. Two categories target known failure modes: walks at night (Low light) and rides on moving platforms (Moving). The ground truth is withheld, and trajectories are scored by the benchmark server against surveyed control points. We made two submissions; the second added the divergence reset and changed three sequences.

The benchmark is far from saturated. Apart from ours, only Aria's own SLAM scores above 35 in any category; established and recent learned systems lose track on the long walks (Table~\ref{tab:slam_lamaria}). Our system ranks second and exceeds the best other entry by 26--46 points in every category. Its trajectories lie within 5\,m of the control points for 96--100\% of the points, against 37--73\% for the best other entries.

\paragraph{Evaluation on public benchmarks.}
Run unchanged on five public benchmarks with different rigs and motion (Table~\ref{tab:slam_ate}), with only scene-scale settings such as the largest accepted loop-closure correction following the environment, the system is on par with or better than the strongest published systems on TUM-VI rooms, Monado, and the Aria datasets, and close to OKVIS2 on EuRoC. Two limitations remain. Very fast rotation (above roughly 500\,$^\circ$/s, where some gyroscopes also saturate) defeats the patch tracker, causing two of the three Monado failures (boxing-game sequences) and the larger errors on the fast VECtor sequences. Loop-closure recall in long featureless halls is weaker than with longer feature tracks: across all 20 TUM-VI sequences, where hall sequences have ground truth only in the start and end room, our median error is 8.1\,cm.

\begin{figure*}[t]
  \centering
  \renewcommand{\arraystretch}{1.1}
  \begin{tabular*}{\linewidth}{@{\extracolsep{\fill}}ccc@{}}
    \includegraphics[width=0.30\linewidth]{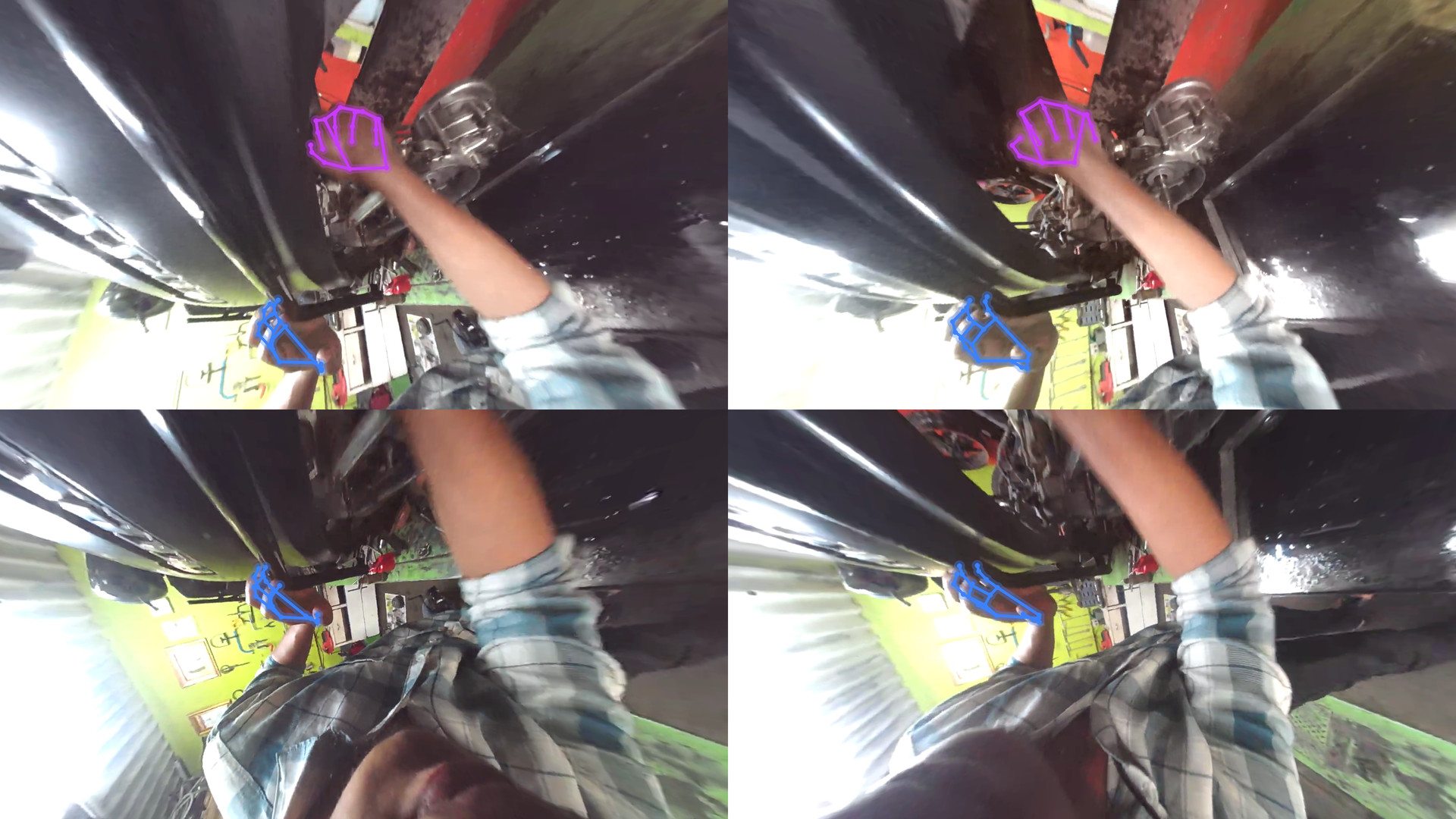} &
    \includegraphics[width=0.30\linewidth]{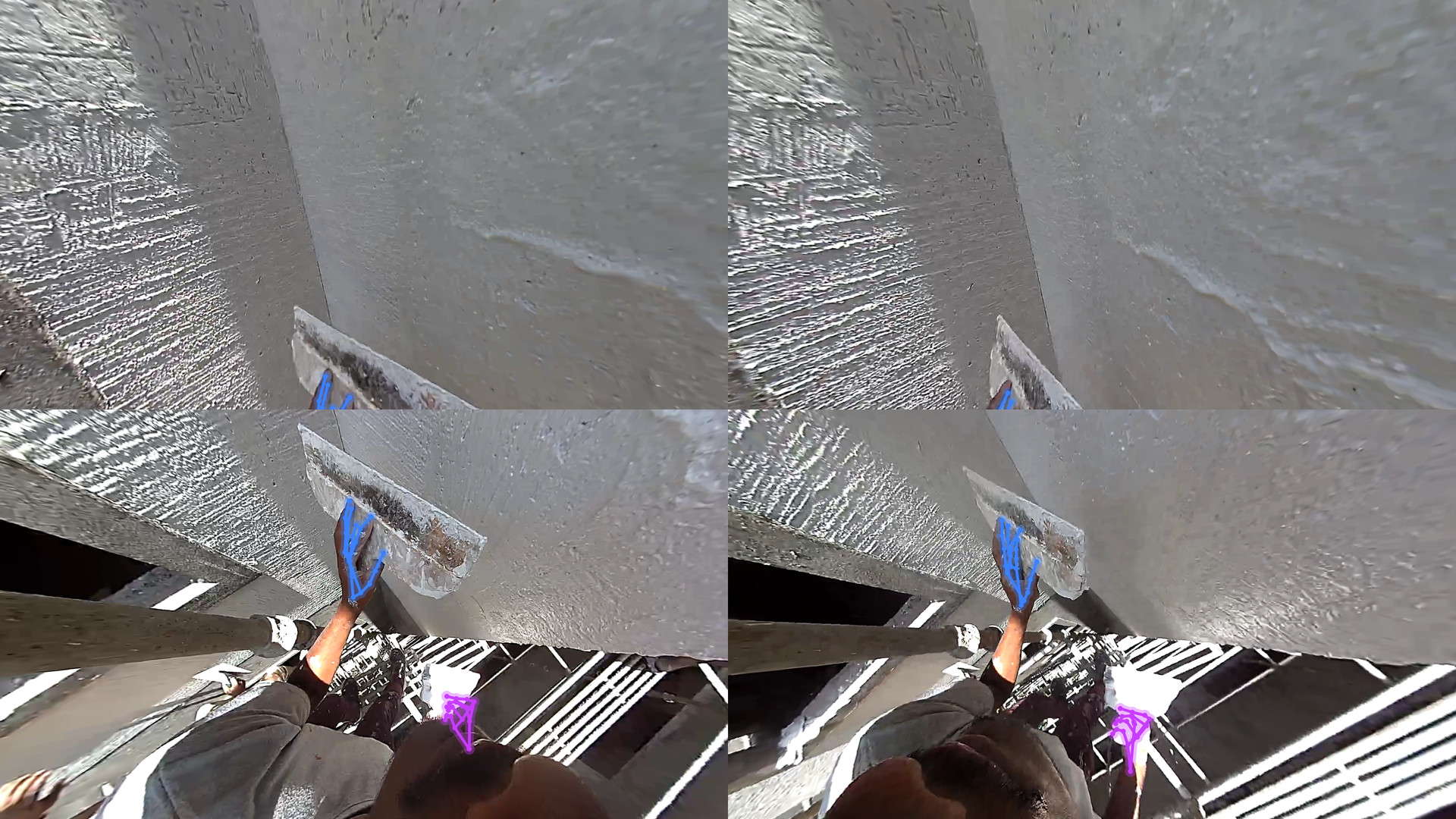} &
    \includegraphics[width=0.30\linewidth]{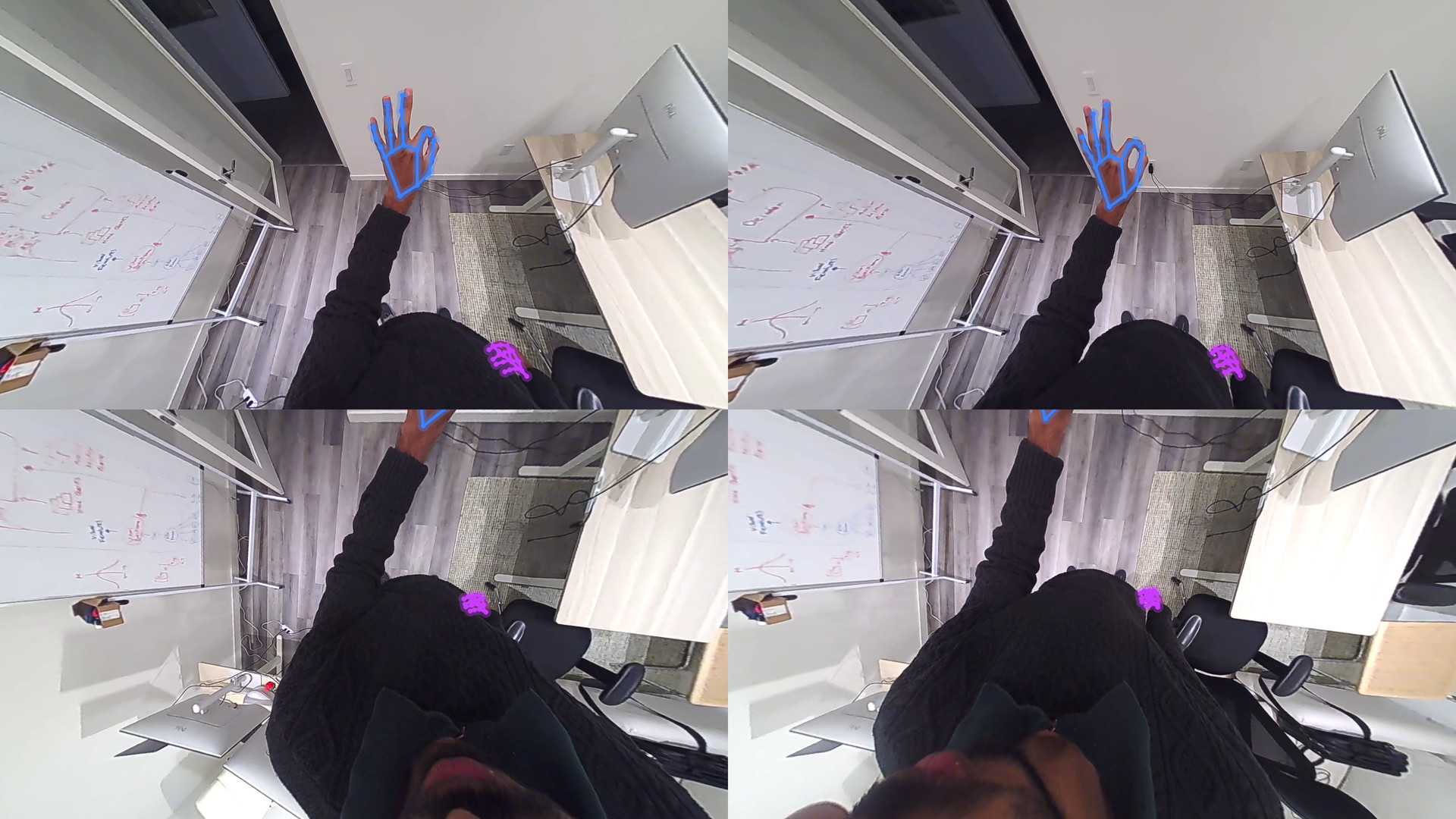} \\
    \small (a) Motion blur &
    \small (b) Visual ambiguity &
    \small (c) Complex hand poses \\[6pt]
    \includegraphics[width=0.30\linewidth]{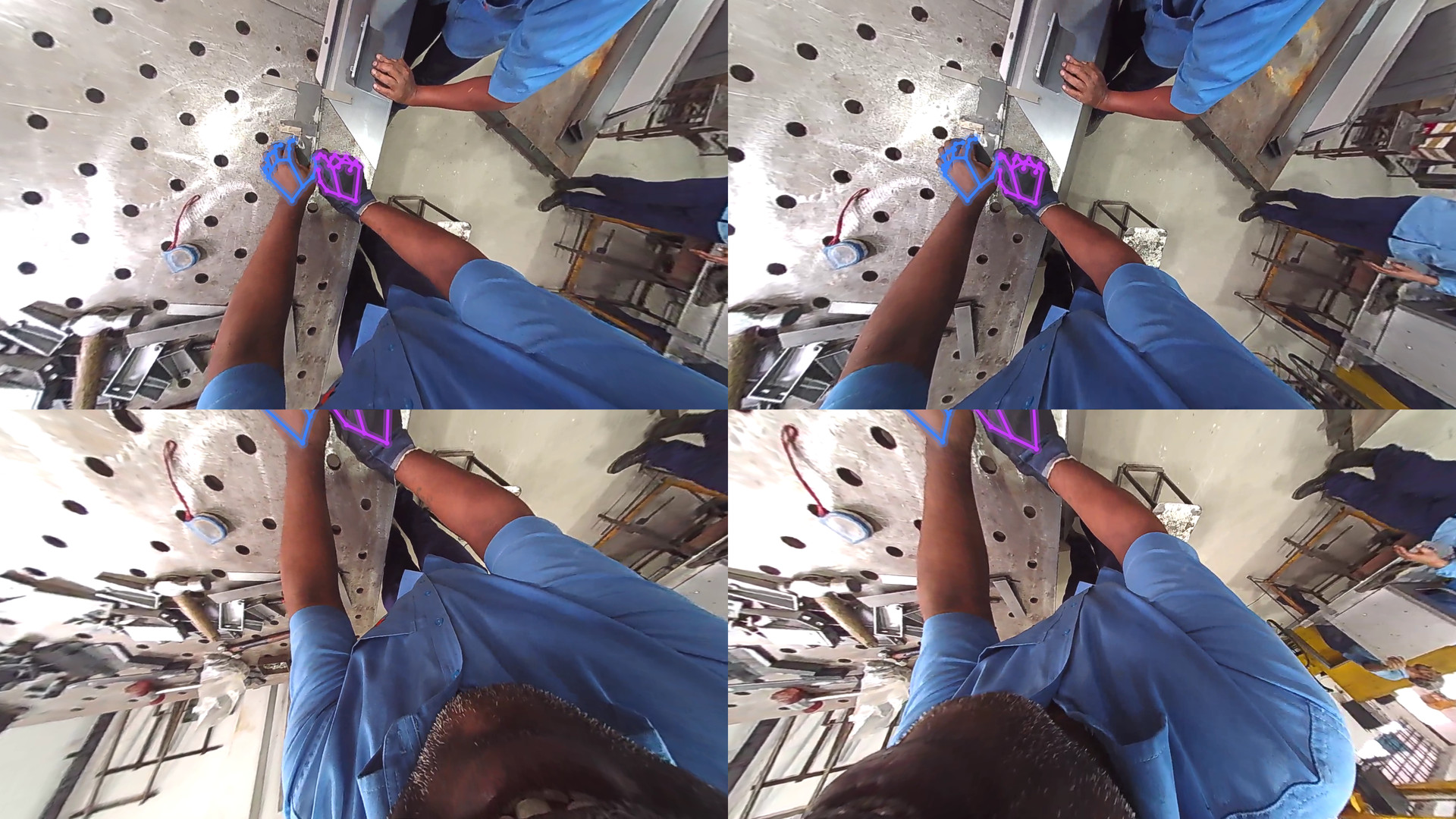} &
    \includegraphics[width=0.30\linewidth]{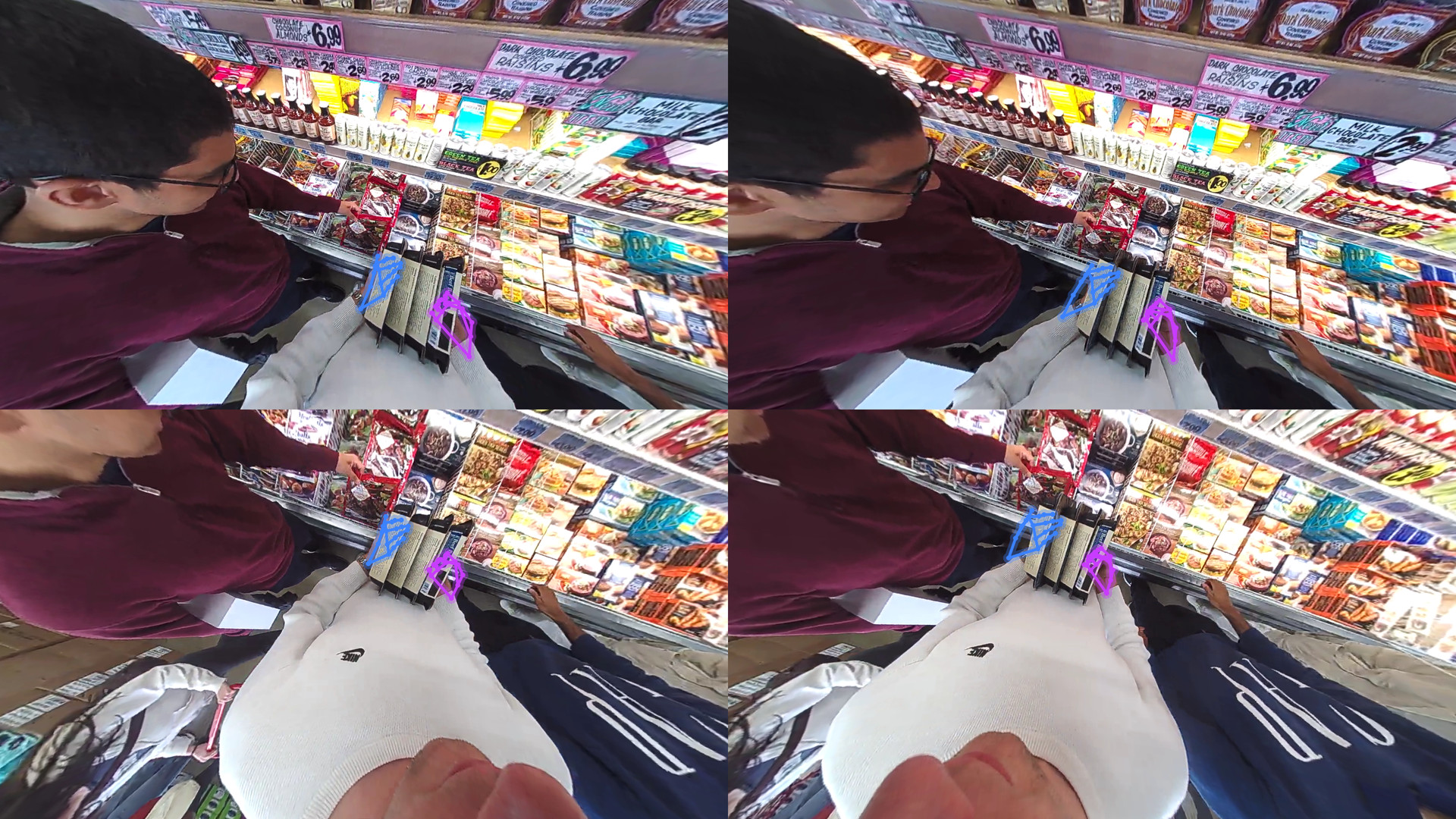} &
    \includegraphics[width=0.30\linewidth]{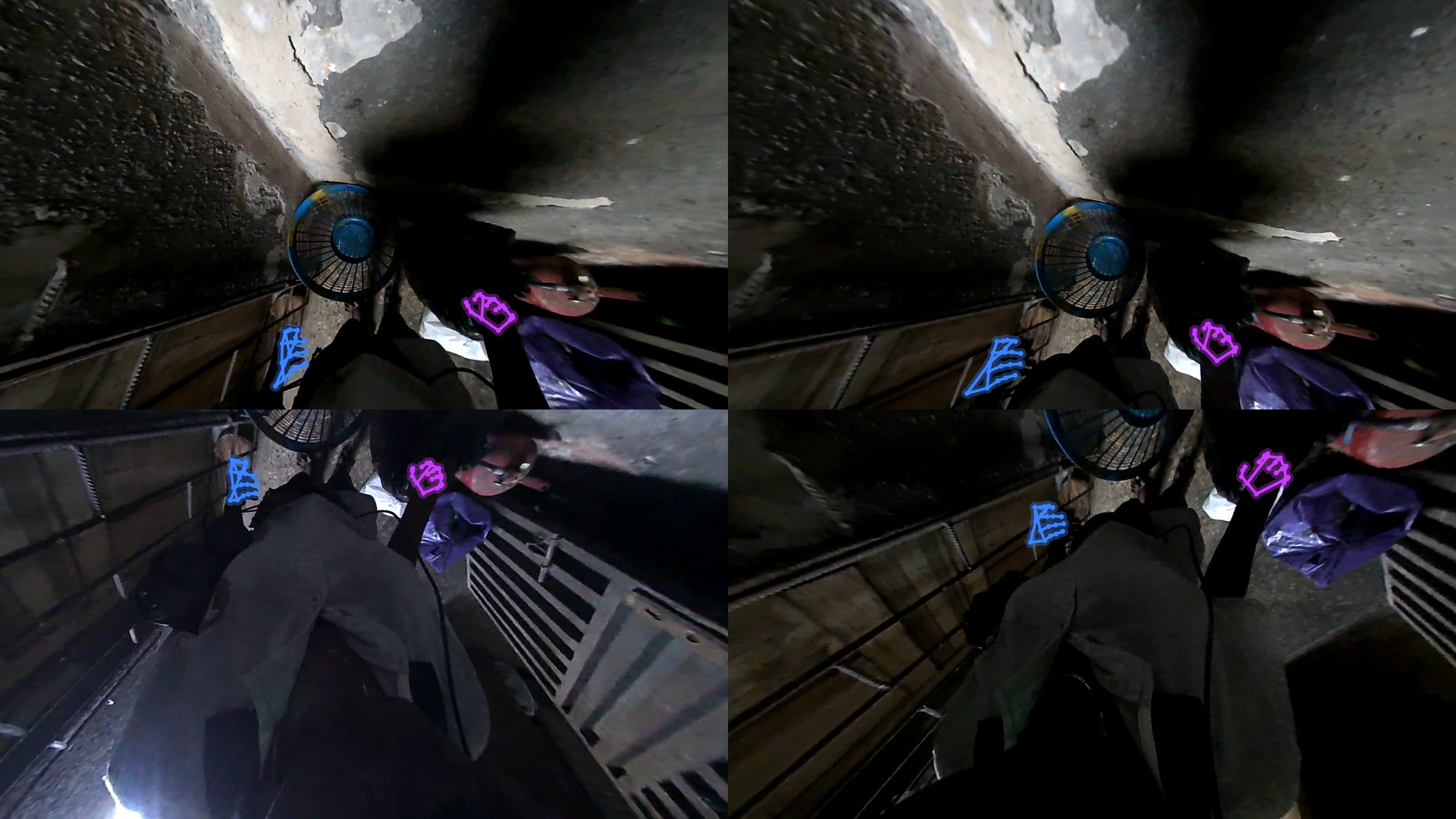} \\
    \small (d) Gloves &
    \small (e) Occlusion &
    \small (f) Lighting \\
  \end{tabular*}
  \caption{Examples from our hand tracking model, which fuses multi-view images into a single hand mesh. Our model's pretraining makes it robust to blur, visual ambiguity, complexity, gloves, occlusion, and lighting. See Figures~\ref{fig:hand_tracking_hard2} and ~\ref{fig:hand_tracking_hard3} for more difficult examples.}
  \label{fig:hand_tracking_hard}
\end{figure*}

\subsection{Depth estimation}
\label{sec:depth}

We estimate metric depth on both stereo pairs with Fast-FoundationStereo (FFS)~\cite{wen2026fastfoundationstereo}, a lightweight model distilled from FoundationStereo (FS)~\cite{wen2025foundationstereo}, and find that model performance is dictated more by calibration than generalization ability.

\paragraph{Evaluating zero-shot transfer to egocentric scenarios.}

To evaluate the ability of these zero-shot models in egocentric settings, we benchmark them on UnrealEgo2~\cite{akada20243dhumanposeperception}, a synthetic head-mounted stereo benchmark with dense ground truth. FFS reaches AbsRel 0.016 on surfaces both cameras see at 0.3--1.0\,m, at the benchmark's 2\,cm quantization limit and on par with FS (Table~\ref{tab:ue2_models}). The remaining error lies on pixels only one camera sees, where there is nothing to match. FS fills these better from context, at over 10$\times$ the inference cost. We conclude that zero-shot models are accurate enough, and that the limiting factor on deployed hardware is online calibration (Section~\ref{sec:online_calibration}).

\paragraph{Evaluating the effect of online calibration.}

To measure the effect of calibration on depth, we record a ChArUco board over seven sessions on one device and take ground truth from the board pose, independent of the stereo extrinsics. Online calibration halves FFS depth error within reach (Table~\ref{tab:charuco}). Across a subset of 150 deployed devices, factory calibration leaves a median front-pair epipolar residual of 3.8\,px, which online calibration reduces to 0.56\,px. Switching from factory to online calibration changes FFS depth within reach by a median 3.6\% per device (11\% at the 90th percentile), moving 31\% of in-reach pixels by more than 5\%.

\subsection{Hand tracking}
\label{sec:hand_tracking}

Hand tracking recovers metric MANO~\cite{romero2017embodied} hand pose in the world frame. Egocentric hand tracking is a 3D estimation problem under unusually hard visual distributions: in-the-wild recordings are dominated by visual and geometric edge cases rather than clean lab imagery, and no single camera sees the hands at all times, so estimates must fuse views of very different appearance under frequent occlusion. As a result, ground truth labels must be generated from in-the-wild recordings, which rules out motion capture as a reliable and scalable method of data collection. Furthermore, we pose hand tracking as a multi-view problem, since monocular hand models are both metrically ambiguous and inconsistent across views. Each view independently regresses shape and depth under a weak-perspective camera, so direct triangulation of monocular outputs produces distorted hands that accumulate many centimeters of error in the local hand frame alone.

We train our hand tracking model on a custom in-the-wild RoboCap dataset, spanning diverse skin tones, gloves, object occlusions, and motion blur. The labels for this dataset are produced by an offline multi-view optimization pipeline without any motion capture. We bootstrap monocular priors~\cite{pavlakos2024hamer, potamias2024wilor} to learn a multi-view hand reconstruction model, then fit a single MANO parameterization via differentiable refinement that optimizes for reprojected appearance and naturalness throughout the trajectory. The hand-tracking model is a feed-forward multi-view ViT that consumes the raw bbox crops and camera parameters to predict a single hand pose across all views, guaranteeing multi-view consistency by construction.


We evaluate three things: the contribution of our pretraining dataset, the
accuracy of the model relative to other multi-view hand estimation models,
and its accuracy on the RoboCap. We report mean per-joint position error
(MPJPE) in millimeters throughout, and detail splits and preprocessing in
Appendix~\ref{sec:eval_setup}.

\paragraph{Evaluating the benefit of the pretraining corpus.}
To isolate the effect of the RoboCap dataset, we train the same architecture on three public egocentric datasets, varying only the
initialization. We select the best cross-dataset validation performance, which evaluates MPJPE on held-out subjects. HOT3D~\cite{banerjee2024hot3d},
UmeTrack~\cite{han2022umetrack} and SHOW3D~\cite{rim2026show3dcapturingscenes3d}
span three headset types with different camera counts, placements and optics,
so the comparison covers rigs unlike the one used for pretraining. None of the
three releases a public test benchmark, so we construct held-out splits that
follow the official protocols as closely as possible; we describe this in
Appendix~\ref{sec:eval_setup}. For each benchmark, we fine-tune our pre-trained model and compare it with an identical model trained from scratch. Pretraining reduces error on all three datasets,
by $2.9$ to $7.5$\,mm ($17$--$40\%$).
\begin{table}[t]
\centering
\begin{tabular}{lcc}
\toprule
& Scratch & Pretrained  \\
\midrule
HOT3D~\cite{banerjee2024hot3d} & 17.41 & \textbf{14.47}  \\
UmeTrack~\cite{han2022umetrack} & 19.00 & \textbf{11.49}  \\
SHOW3D~\cite{rim2026show3dcapturingscenes3d} & 19.34 & \textbf{13.68} \\
\bottomrule
\end{tabular}
\caption{Best validation MPJPE (mm) on held-out subjects across datasets, varying only initialization. More details in Appendix~\ref{sec:appx_hand_tracking}.}
\label{tab:ego_ablation}
\end{table}

\paragraph{Comparison to multi-view reconstruction.}
We compare against POEMv2~\cite{poemv2}, a state-of-the-art multi-view hand pose estimation model, on OakInk~\cite{yang2022oakinklargescaleknowledgerepository}, DexYCB~\cite{chao2021dexycbbenchmarkcapturinghand} and HO3D~\cite{hampali2021ho3dv3improvingaccuracyhandobject}. As our model is
egocentric and pretrained on a single rig, these exocentric evaluation benchmarks are out of
distribution in both viewpoint and camera configuration. We finetune our hand tracking ViT for 20k steps. While we surpass the baseline on DexYCB and OakInk, our performance drops on HO3D. The drop in performance on HO3D likely happened because some test sequences drop 2/5 cameras for evaluating cross-rig generalization. Our model, which was trained on a single consistent rig, struggles to generalize to views/rigs different from the RoboCap without sufficient finetuning.
\begin{table}[t]
\centering
\begin{tabular}{lcc}
\toprule
& POEMv2~\cite{poemv2} & Ours \\
\midrule
DexYCB-Mv & 6.69 & \textbf{5.22} \\
OakInk-Mv & 8.34 & \textbf{7.25} \\
HO3D-Mv & \textbf{7.72} & 13.01 \\
\bottomrule
\end{tabular}
\caption{MPJPE (mm) on third-person multi-view benchmarks.}
\label{tab:multiview}
\end{table}

\paragraph{Evaluating performance on RoboCap.}
Our final evaluation is on our own proprietary sim benchmark, rendered with RoboCap camera extrinsics and intrinsics; we detail its construction in Appendix~\ref{sec:appendix_synthetic_data}. We achieve sub-centimeter accuracy ($9.2$\,mm). In general, we do not believe that sim and motion capture data represents the difficulty of in-the-wild hand tracking, but still construct an internal benchmark to test the upper-bound performance of our models.

\section{Limitations}
\label{sec:limitations}

Our hardware makes a specific bet. Designing the sensing suite backward from the requirements of metric 3D estimation buys precision at manipulation range, but it also fixes the geometry. The working volume is set by the stereo baselines and the lateral coverage, and reaching motions that leave every frustum are not recoverable. The device also assumes a tethered battery bank, which is acceptable for deliberate collection and not for ambient all-day wear. We regard the latter as a form-factor tradeoff rather than a permanent one.

Our SLAM matches or exceeds the published systems on every benchmark we report, but not uniformly. It struggles on long, self-similar loops, where too few revisits are recognized to correct the drift accumulated around them; this is a limit of place-recognition recall rather than of the estimator. Aria's own MPS pipeline is strong in exactly this regime, and on ADT and HOT3D we are scored against it, so the sub-centimeter figures there should be read as agreement with a pseudo-reference rather than as accuracy, and they bound what those datasets can demonstrate about either system. Finally, the causal estimate available during capture is about twice as inaccurate as the offline trajectory, since the refinement that produces the reported result runs over the whole session once capture has ended; applications that need the final accuracy online would not be served by this system today.

Hand tracking evaluation remains the weakest link in this setting, and not only for us. No public dataset provides what egocentric manipulation calls for at once: short-baseline rectified stereo, a head-worn viewpoint, and dense independent ground truth under visually diverse edge cases. We therefore design our benchmarks to measure the quality of our method transfer indirectly against ground truth on other rigs, as is standard in 3D computer vision literature.

Finally, this report establishes precision and not its downstream effect on policy performance. We argue in Section~\ref{sec:intro} that label error enters a policy's training distribution as noise, and prior work supports the link~\cite{liu2025egozero, kareer2024egomimic, qiu2025humanoidpolicy}, but we do not measure here how much downstream performance a centimeter buys. That experiment requires training policies at scale on data this stack produces, and is the subject of ongoing work.

\section{Conclusion}
\label{sec:conclusion}

Egocentric data is one of the most scalable data sources available to robot learning. However, collecting it at the precision and scale that policy learning requires has demanded hardware built for 3D estimation and algorithms robust enough to run unattended on in-the-wild footage, and neither has been something a team could buy or call. We built both: RoboCap, a 250\,g six-camera dual-IMU hat whose sensing suite is specified by what metric estimation requires, and the Grounded API, a suite of 3D algorithms that are tuned for it and demonstrate general state-of-the-art 3D capabilities when transferred to other rigs.

The result we find most consequential is not any single benchmark number but what bootstrapping 3D labels for in-the-wild data implies about scale. Specifically, every hand pose corpus derives ground truth from a motion-capture installation or a closed on-device tracker, which confines the diversity and scalability of data collection. Instead, we show that precise and robust SLAM, depth estimation, and hand tracking do not require motion capture, thereby enabling our hand tracking models to adapt better than models trained only on motion capture data.

Both the device and the API are available today. We expect the field to find failure modes in them that we have not, and hope to push the envelope of egocentric data, 3D estimation, and robot learning.

\clearpage
\setcounter{page}{1}
\maketitlesupplementary

\appendix

\begin{strip}
    \centering

    \includegraphics[width=0.9\linewidth]{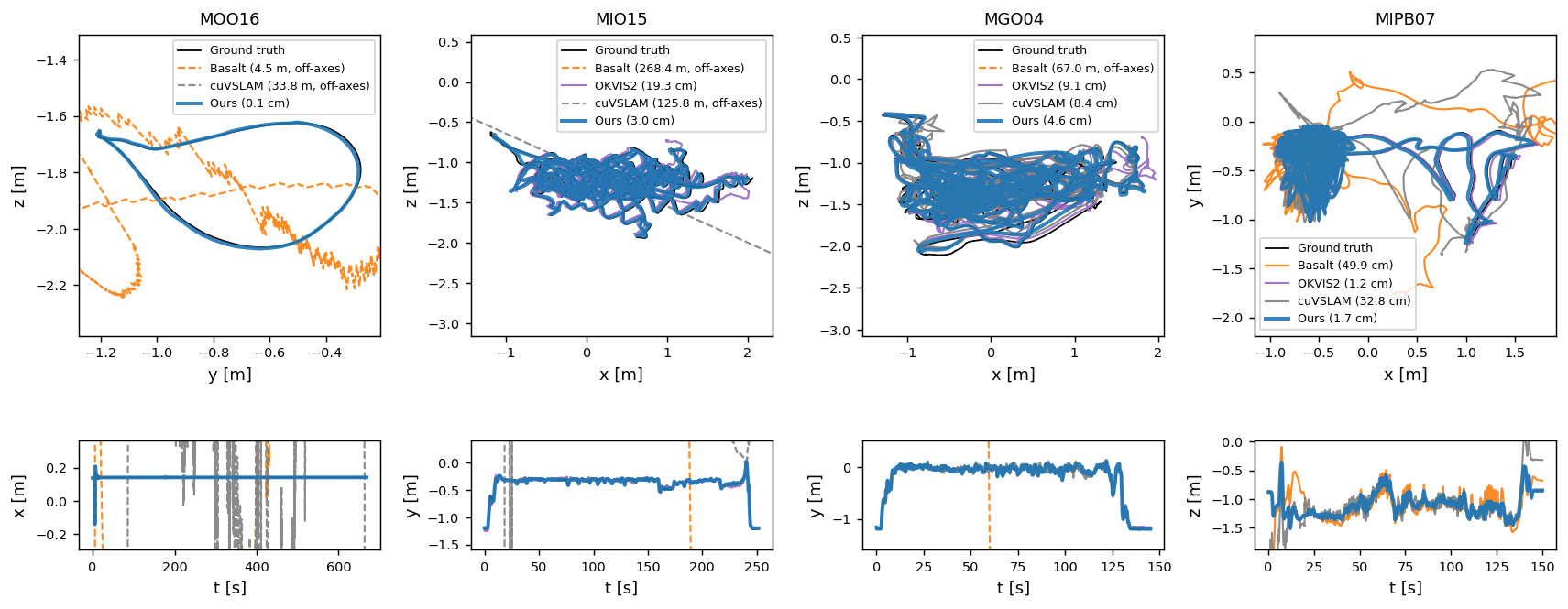}
    \includegraphics[width=0.9\linewidth]{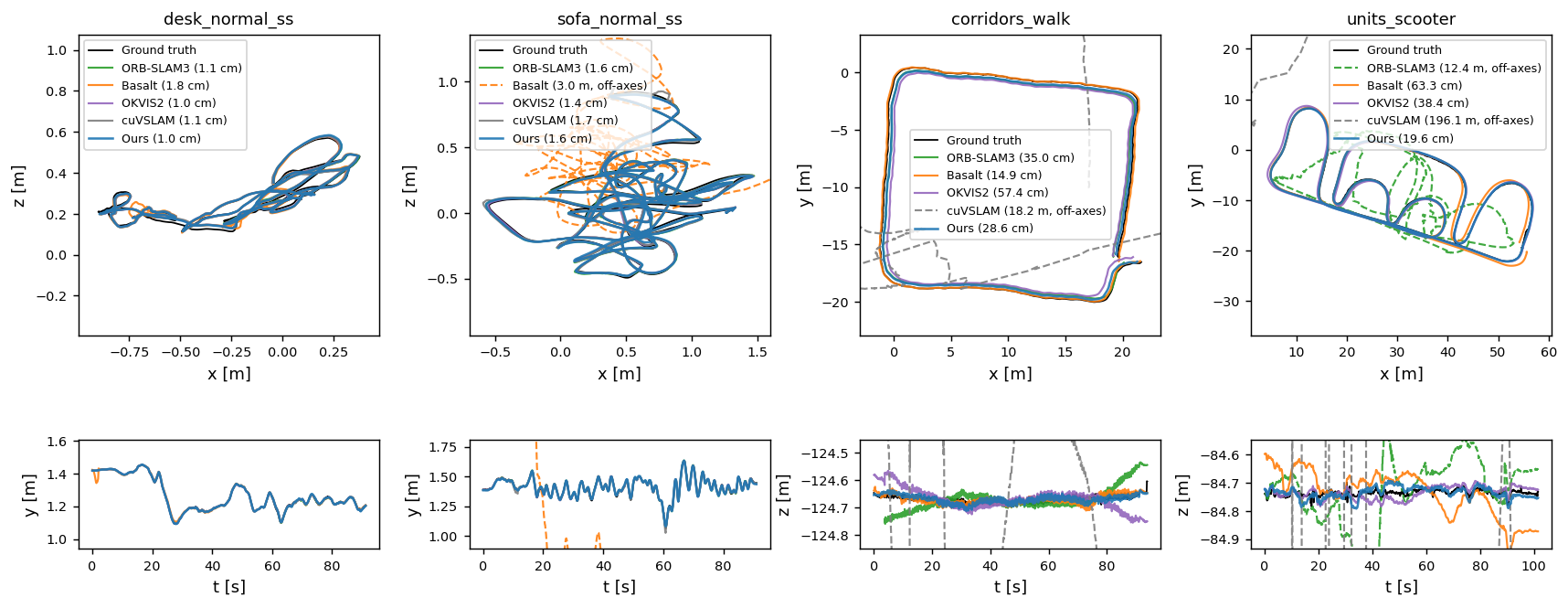}
    \captionsetup{hypcap=false}
    \captionof{figure}{Overlays of SLAM trajectories on Monado (top) and VECtor (bottom).}
    \captionsetup{hypcap=true}
    \label{fig:slam_trajectories}

    \vspace{16pt}

    \begin{tabular}{lccccc}
    \toprule
    Benchmark & ORB-SLAM3~\cite{campos2021orbslam3} & Basalt~\cite{usenko2020basalt} & OKVIS2~\cite{leutenegger2022okvis2} & cuVSLAM~\cite{korovko2025cuvslam} & Ours \\
    \midrule
    EuRoC~\cite{burri2016euroc}  & 3.5 / 3.3 & 5.8 / 5.5 & \textbf{3.1} / \textbf{2.3} & 6.6 / 8.4$^{(4)}$ & \textbf{3.1} / \textbf{2.3} \\
    TUM-VI (rooms)~\cite{schubert2018tumvi} & 0.9 / 0.9 & 9.2 / 6.3 & 0.9 / 0.9 & 2.0 / 1.7 & \textbf{0.7} / \textbf{0.6} \\
    VECtor~\cite{gao2022vector} & \textbf{14.7} / 23.0$^{(5)}$ & 28.0 / 71.1$^{(7)}$ & 25.7 / 16.8$^{(3)}$ & 16.5 / 36.0$^{(7)}$ & 22.8 / \textbf{13.0}$^{(2)}$ \\
    Monado~\cite{monado2023}$^{*}$ & -- / 2.5 & -- / 11.2 & -- / \textbf{2.2} & 11.5 / 11.5$^{(20)}$ & \textbf{4.1} / \textbf{2.2} \\
    ADT~\cite{pan2023adt}$^{\dagger}$    & 2.0 / 1.8$^{(3)}$ & 11.2 / 6.5$^{(3)}$ & 5.4 / 3.0$^{(4)}$ & 15.9 / 6.9$^{(12)}$ & \textbf{1.1} / \textbf{0.9} \\
    HOT3D~\cite{banerjee2024hot3d}$^{\dagger}$  & 1.3 / $\infty$$^{(18)}$ & 4.5 / 4.7$^{(4)}$ & 1.7 / 1.7$^{(6)}$ & 46.9 / 55.4$^{(6)}$ & \textbf{0.4} / \textbf{0.3} \\
    \bottomrule
    \end{tabular}
    \caption{Final (offline) ATE RMSE in cm, mean / median over sequences. A failure or an ATE $>$ 1\,m, denoted with $^{(n)}$, is excluded from the mean, $\infty$ in the median. We use each paper's own numbers where applicable, otherwise we run its released code with the published configuration on the same inputs. VECtor counts \texttt{robot\_fast} as a failure for every method. Monado includes the two long free-movement sessions (64 sequences). $^{*}$Monado baselines are the non-causal aggregate medians reported by the dataset authors, and means are not published. $^{\dagger}$Aria datasets with pseudo-ground truth from Aria MPS.}
    \label{tab:slam_ate}

    \vspace{16pt}

    \includegraphics[width=\linewidth]{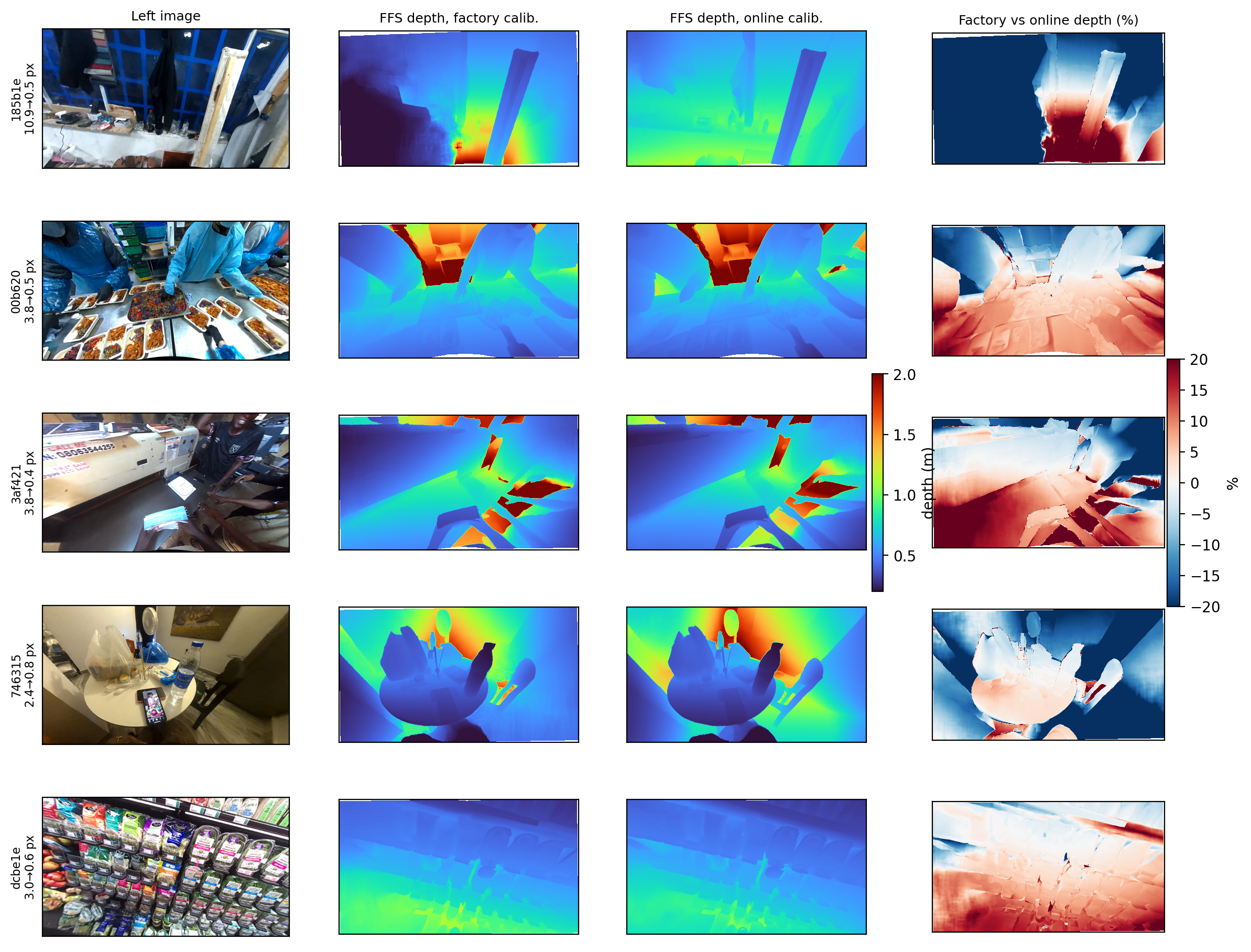}
    \captionsetup{hypcap=false}
    \captionof{figure}{More visualizations of FFS depth with factory vs. online calibration, and the depth change between them.}
    \captionsetup{hypcap=true}
    \label{fig:depth_calib2}

    \vspace{16pt}

    \includegraphics[width=0.30\linewidth]{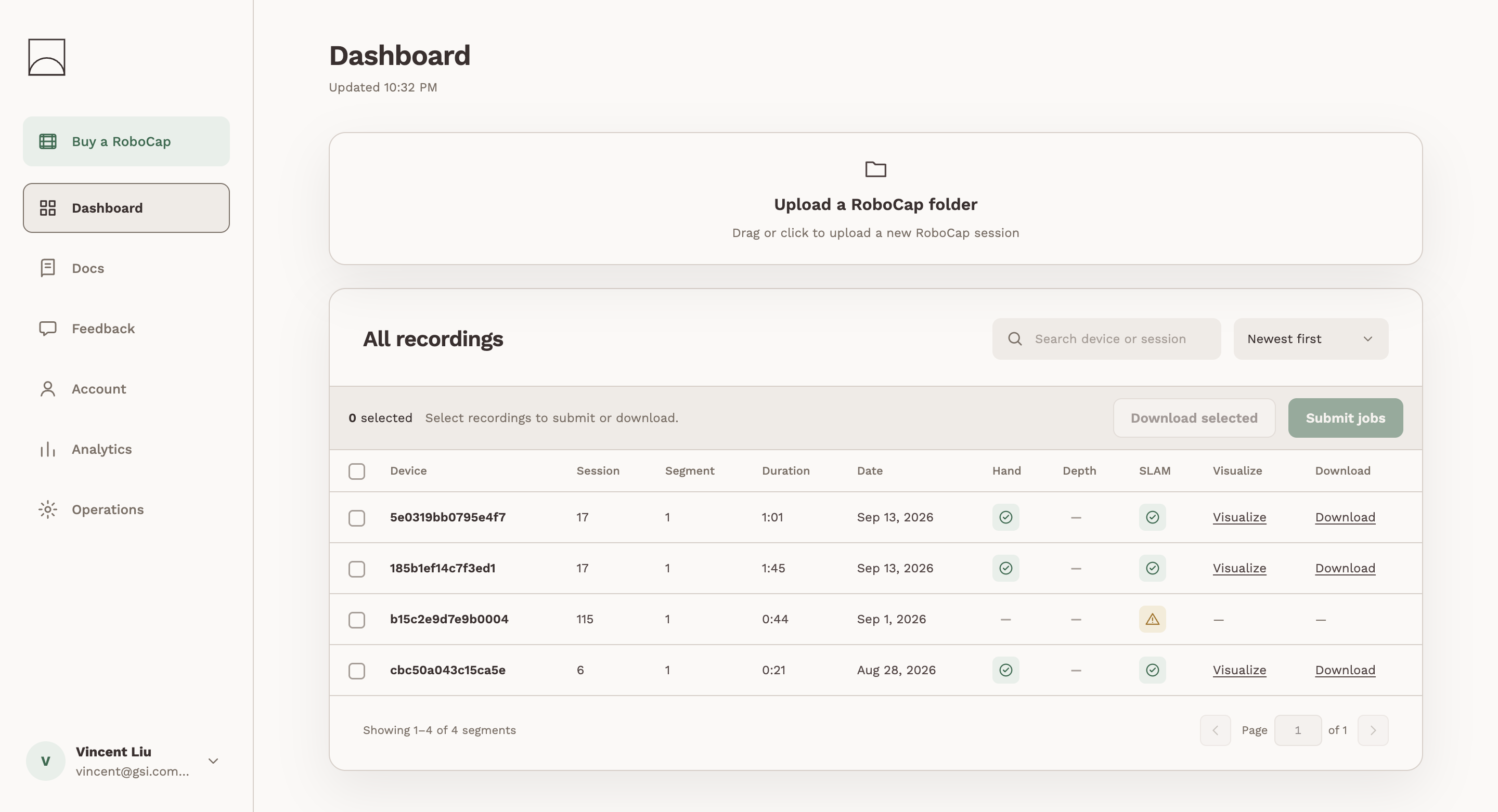}\hfill
    \includegraphics[width=0.30\linewidth]{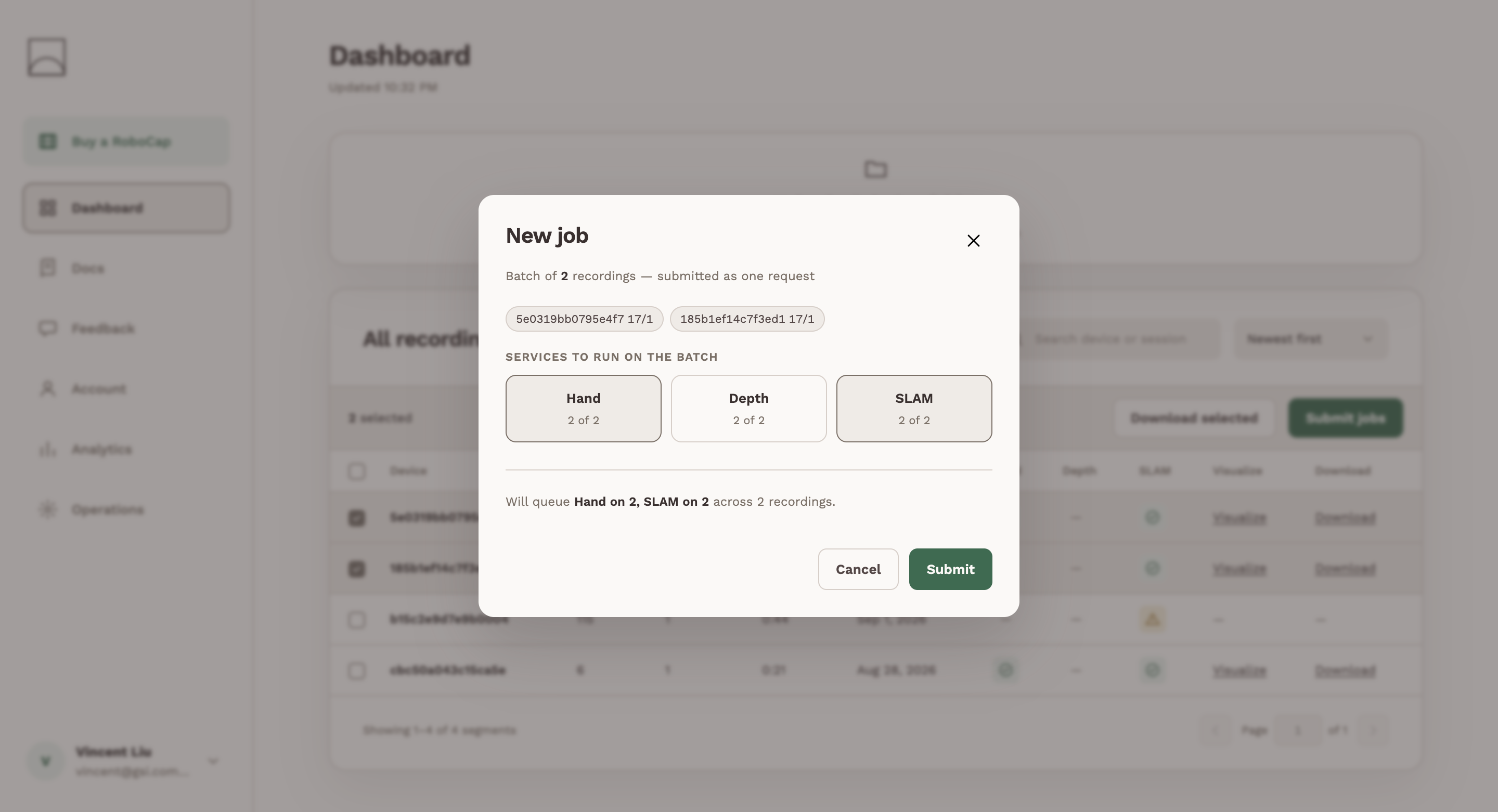}\hfill
    \includegraphics[width=0.30\linewidth]{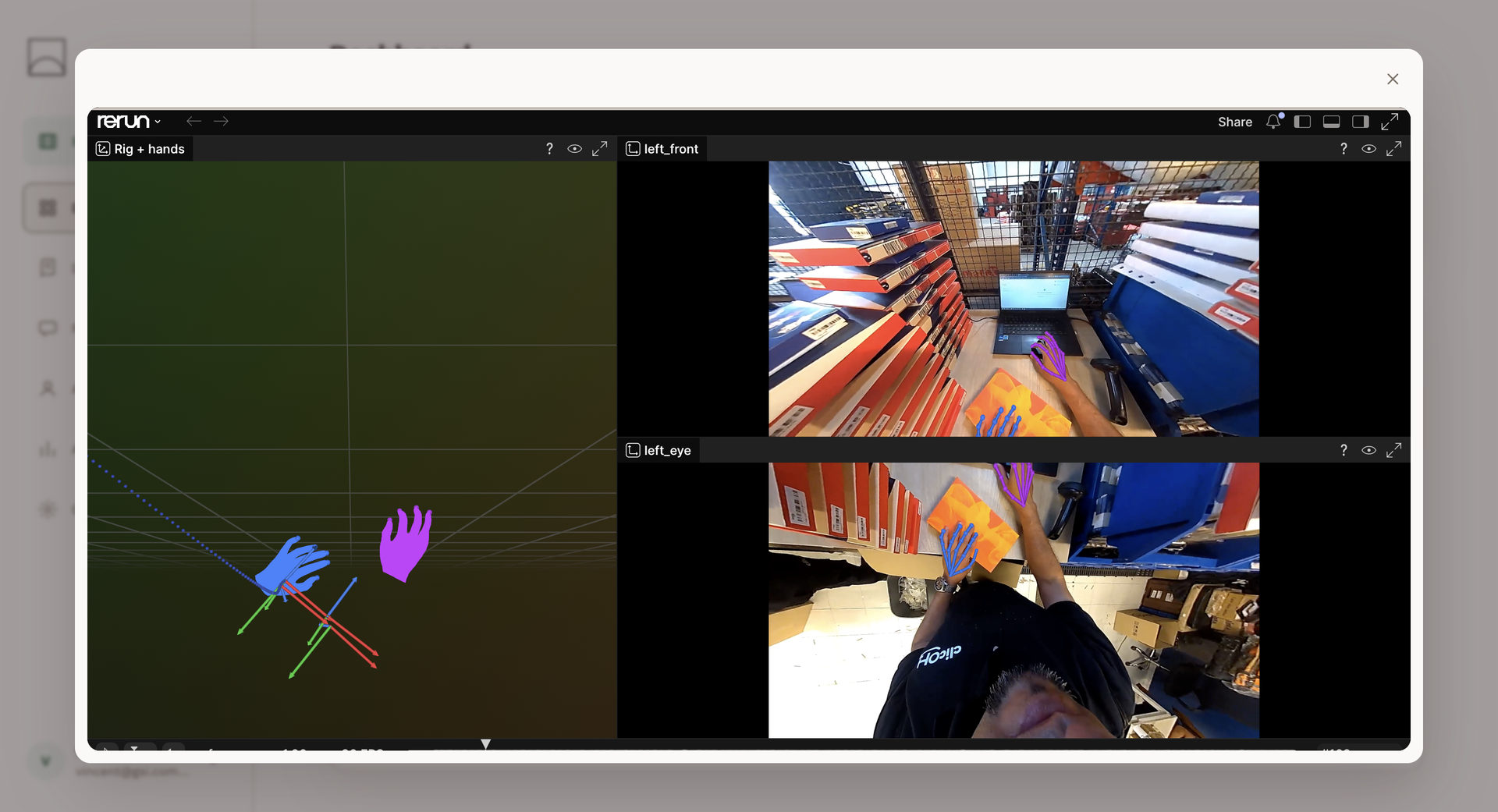}
    \captionsetup{hypcap=false}
    \captionof{figure}{The Grounded API, which allows users to upload RoboCap sessions and submit jobs directly from a dashboard. The dashboard also embeds a 3D visualizer to view hand tracking and SLAM outputs. The CLI supports batch uploads, submissions, and downloads.}
    \captionsetup{hypcap=true}
\end{strip}

\section{Hand tracking}
\label{sec:appx_hand_tracking}

\begin{figure*}[t]
  \centering
  \renewcommand{\arraystretch}{1.1}
  \begin{tabular*}{\linewidth}{@{\extracolsep{\fill}}cc@{}}
    \includegraphics[width=0.46\linewidth]{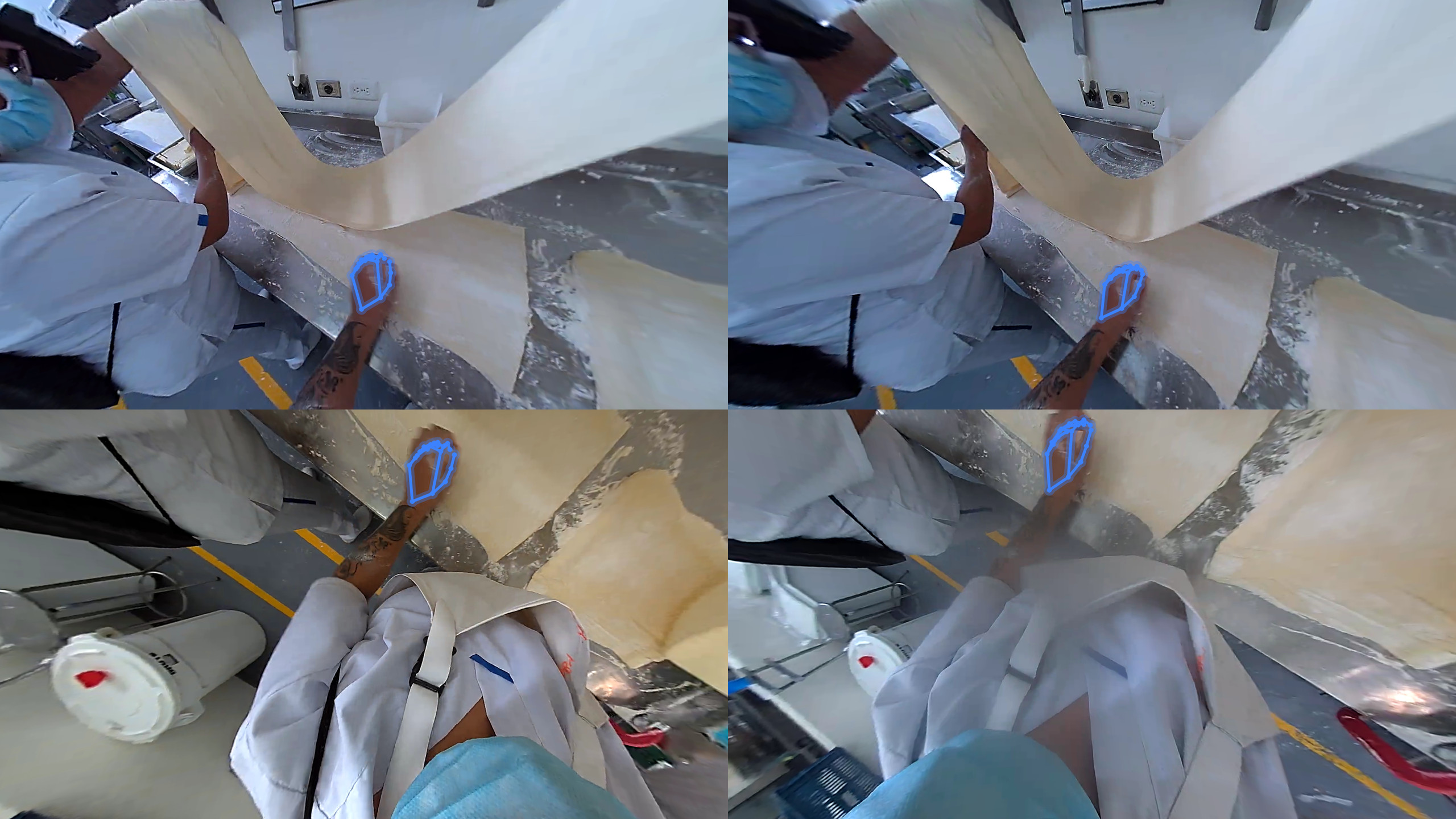} &
    \includegraphics[width=0.46\linewidth]{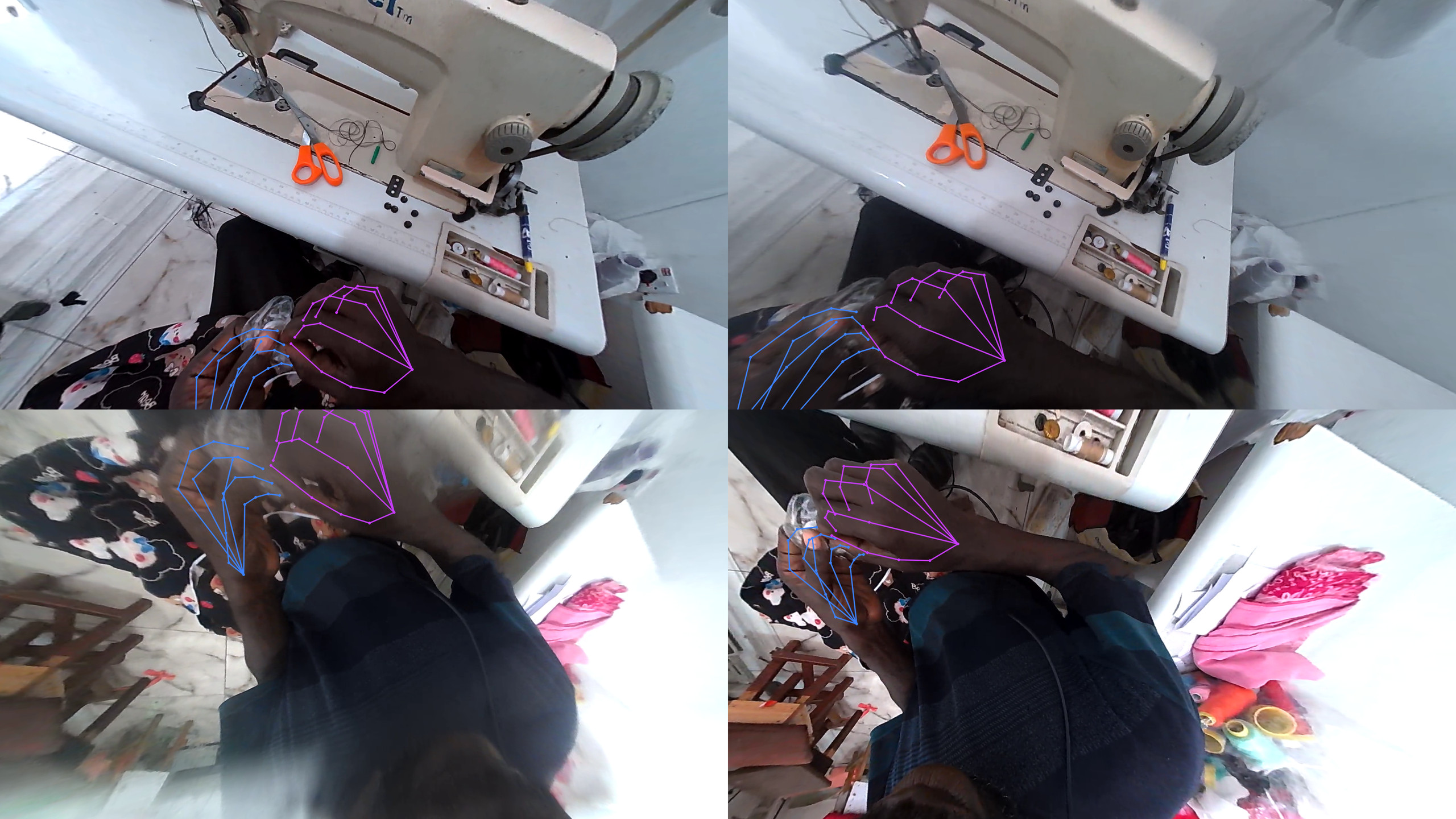} \\
    \small (a) Motion blur &
    \small (b) Lens fog \\[6pt]
    \includegraphics[width=0.46\linewidth]{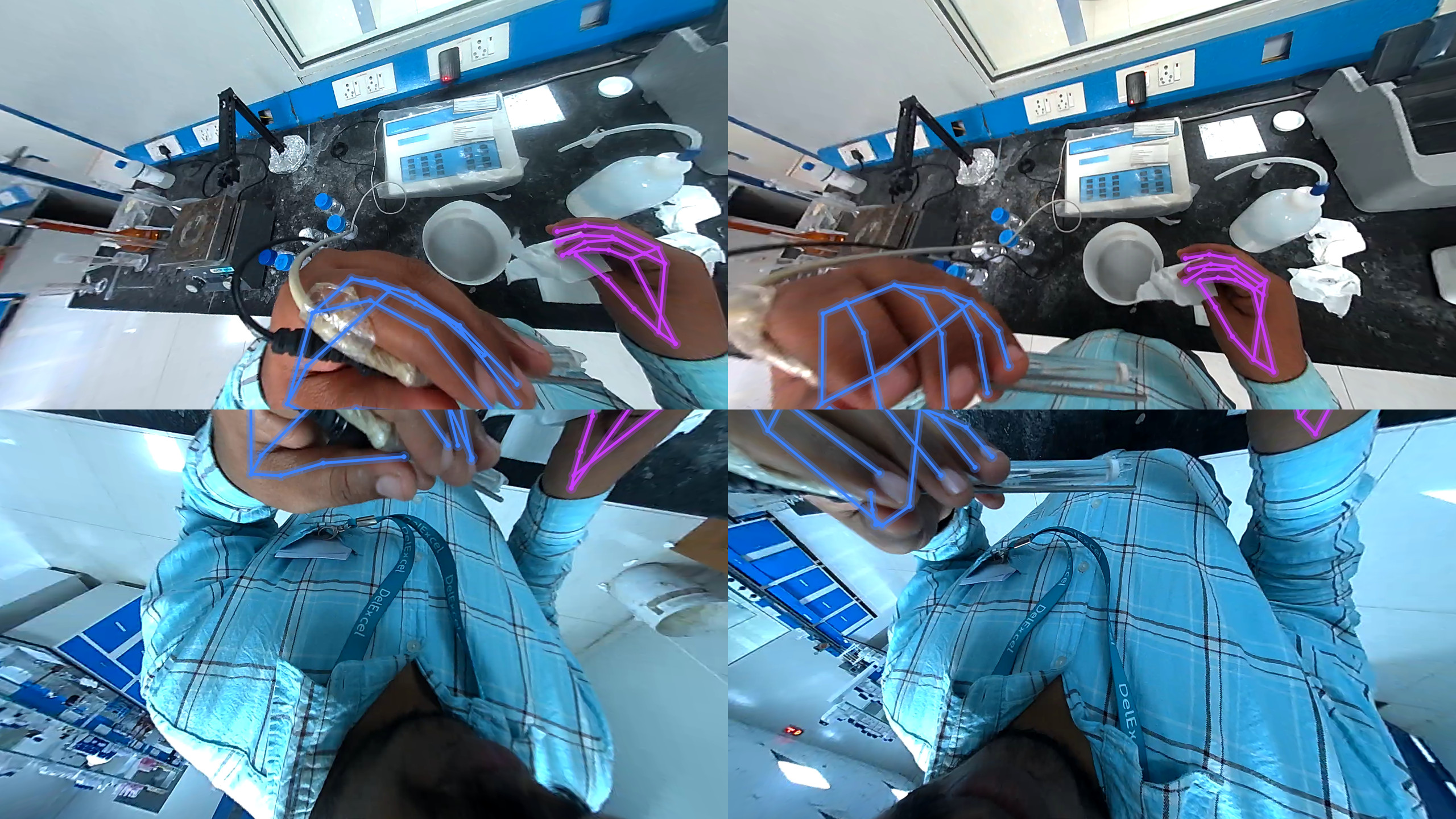} &
    \includegraphics[width=0.46\linewidth]{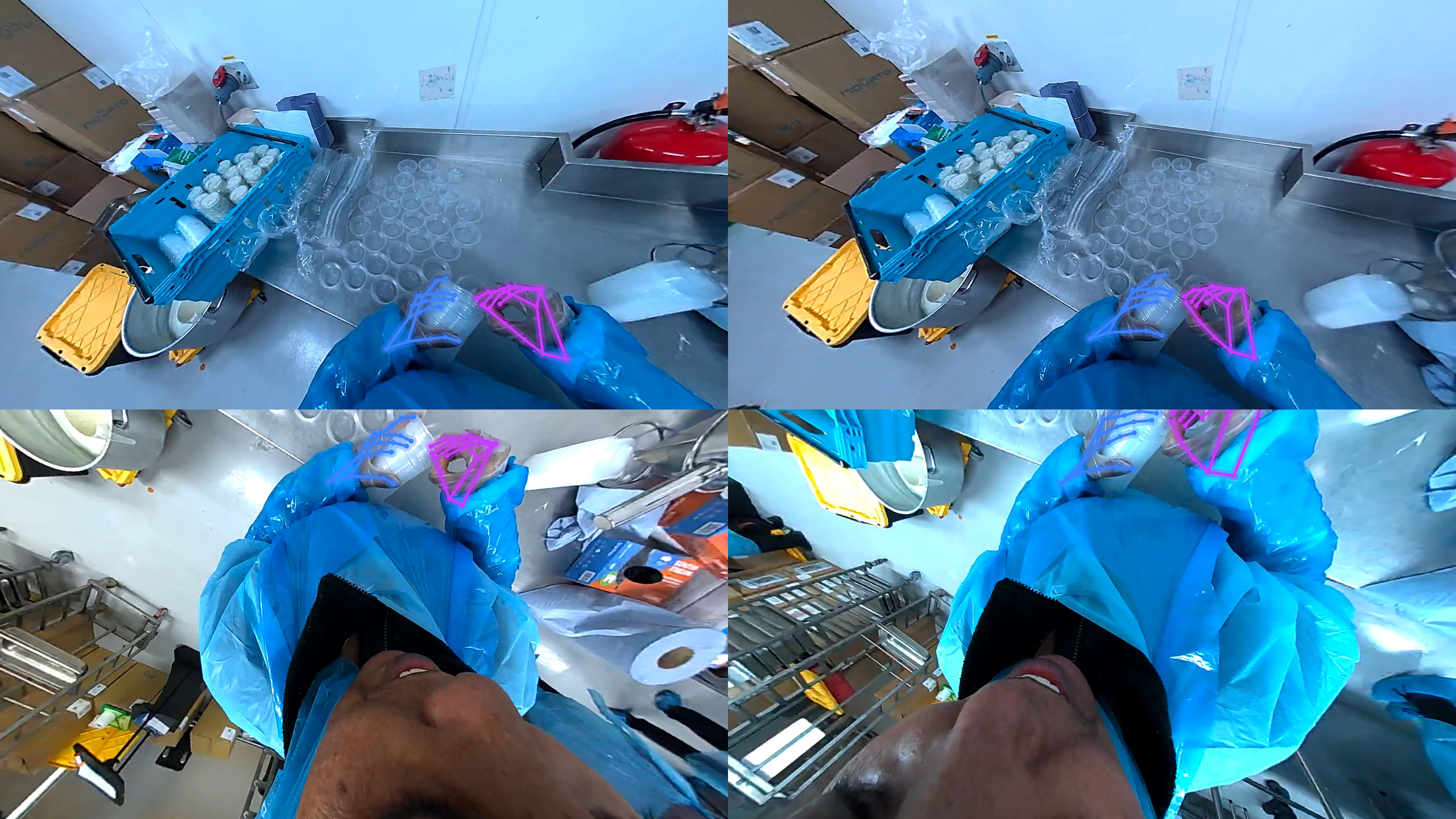} \\
    \small (c) Hand close to camera &
    \small (d) Transparent gloves \\
  \end{tabular*}
  \caption{Robustness to degraded imaging and extreme viewpoints. Each panel
  shows the four camera views of a single frame with the predicted mesh
  overlaid. The model remains multi-view consistent when the hand is blurred by
  fast motion, when the lens is fogged, when the hand is close enough to the
  camera to fill the frame, and when the hand is seen through a transparent
  glove.}
  \label{fig:hand_tracking_hard2}
\end{figure*}

\begin{figure*}[t]
  \centering
  \renewcommand{\arraystretch}{1.1}
  \begin{tabular*}{\linewidth}{@{\extracolsep{\fill}}cc@{}}
    \includegraphics[width=0.46\linewidth]{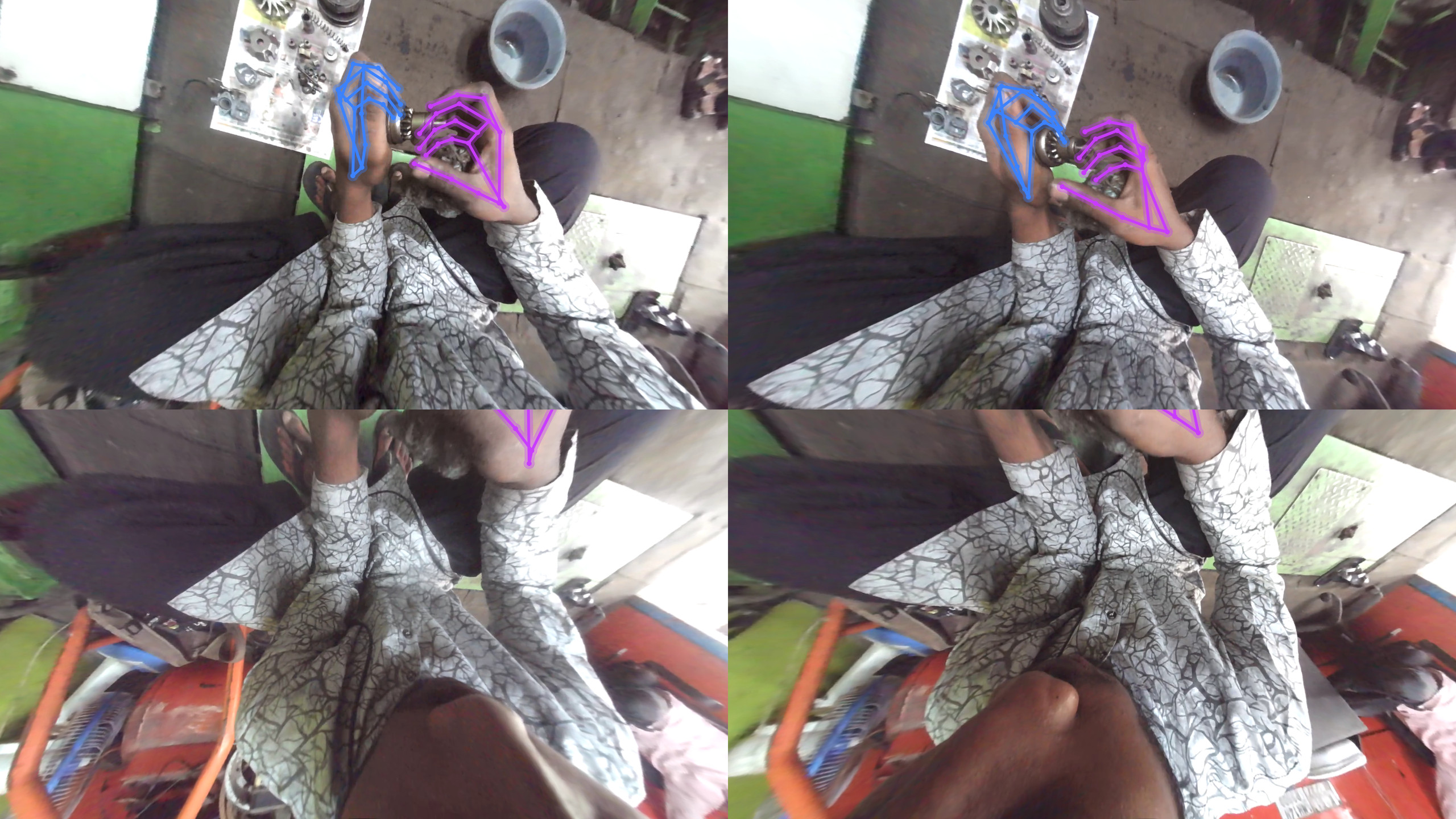} &
    \includegraphics[width=0.46\linewidth]{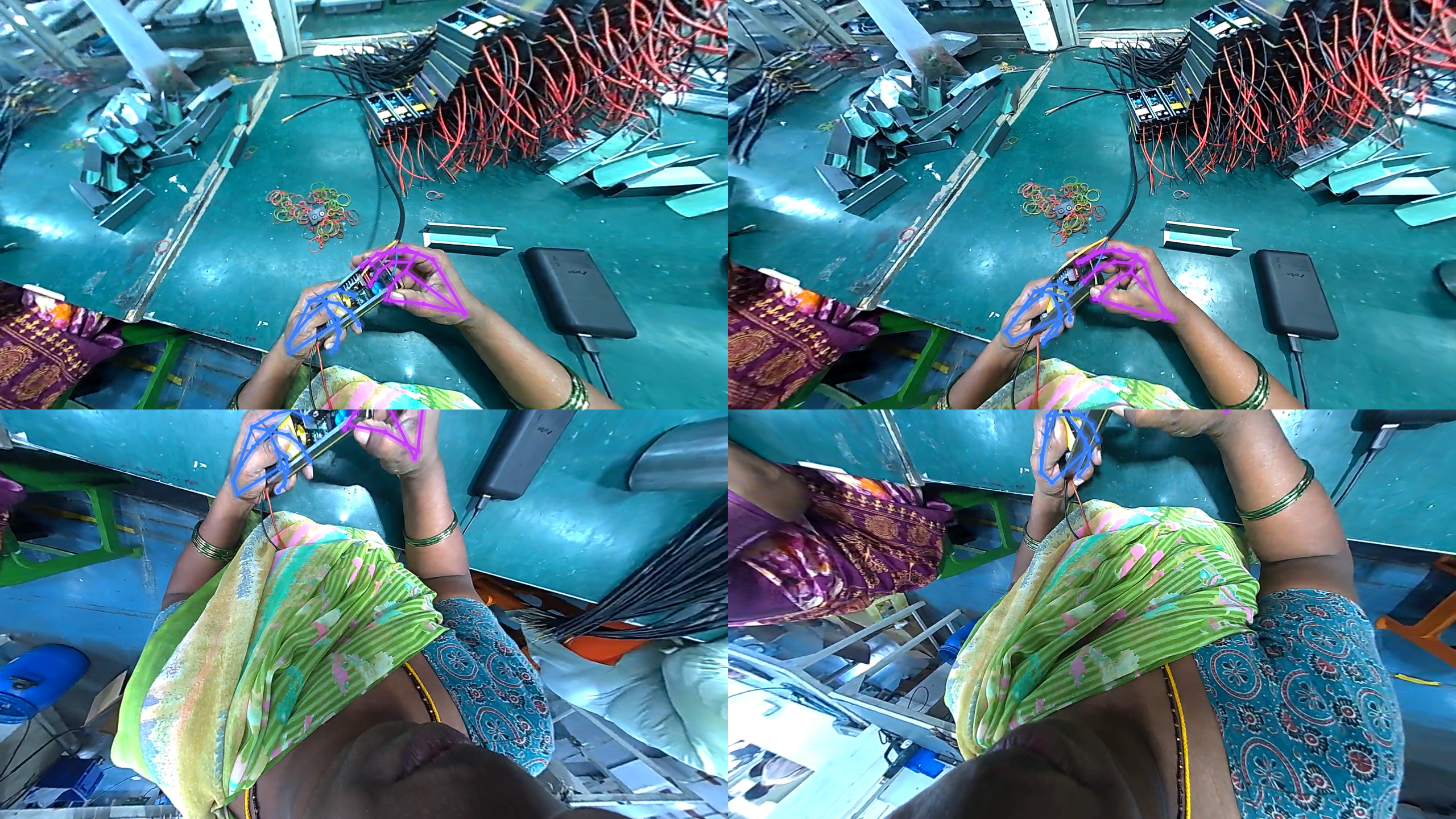} \\
    \small (a) Small objects &
    \small (b) Small objects \\[6pt]
    \includegraphics[width=0.46\linewidth]{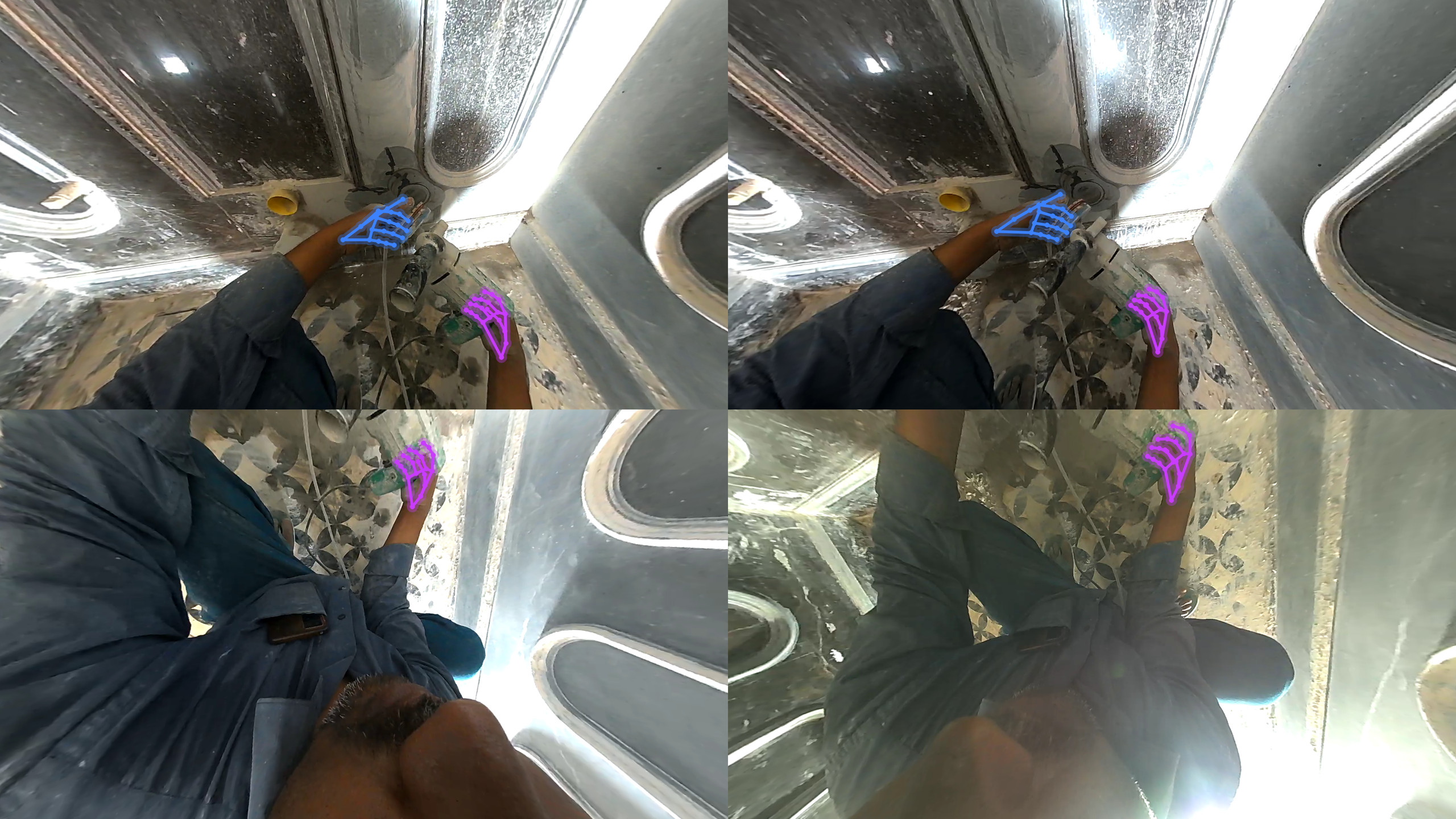} &
    \includegraphics[width=0.46\linewidth]{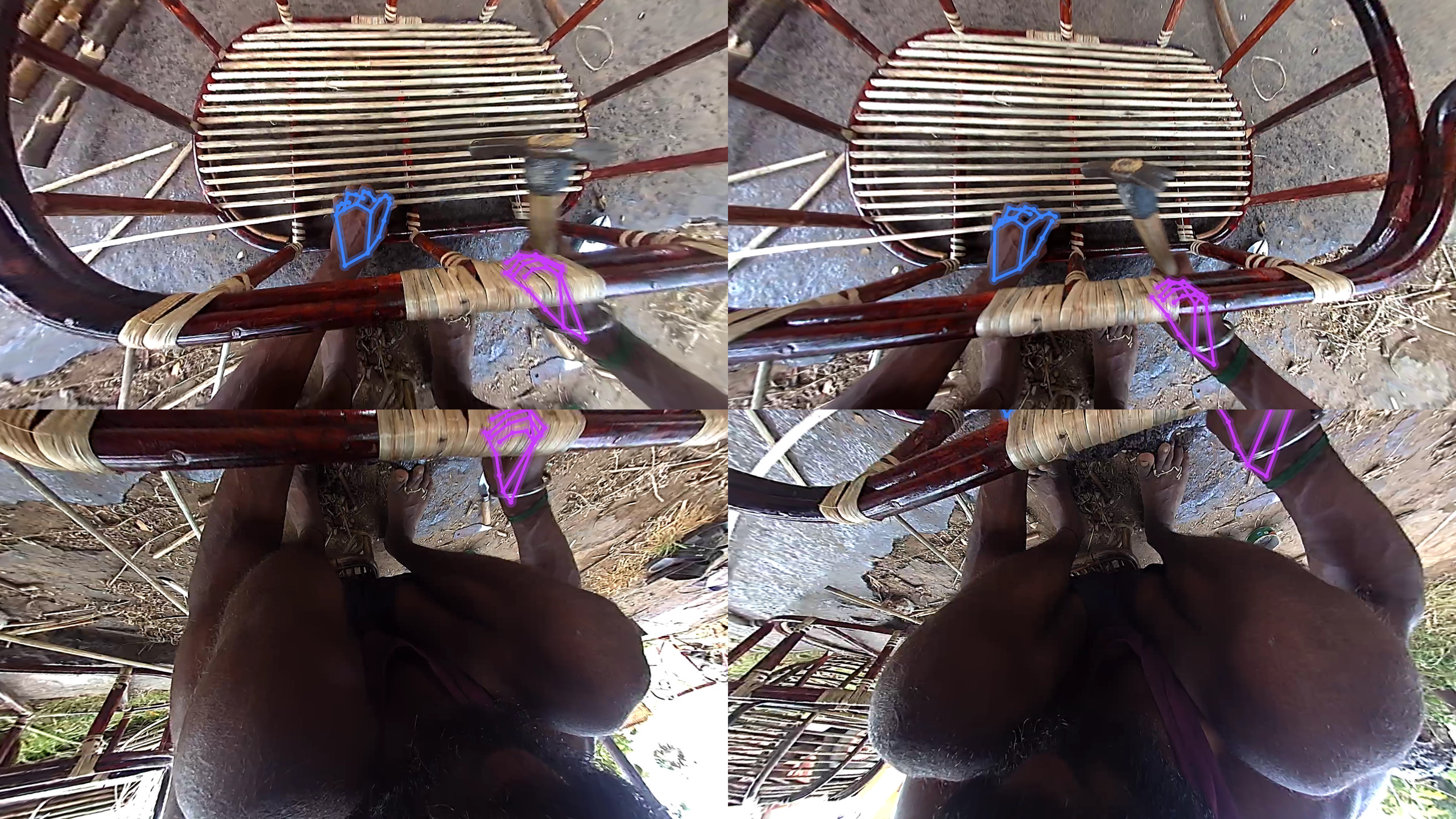} \\
    \small (c) Tool use &
    \small (d) Tool use (with occlusion) \\[6pt]
    \includegraphics[width=0.46\linewidth]{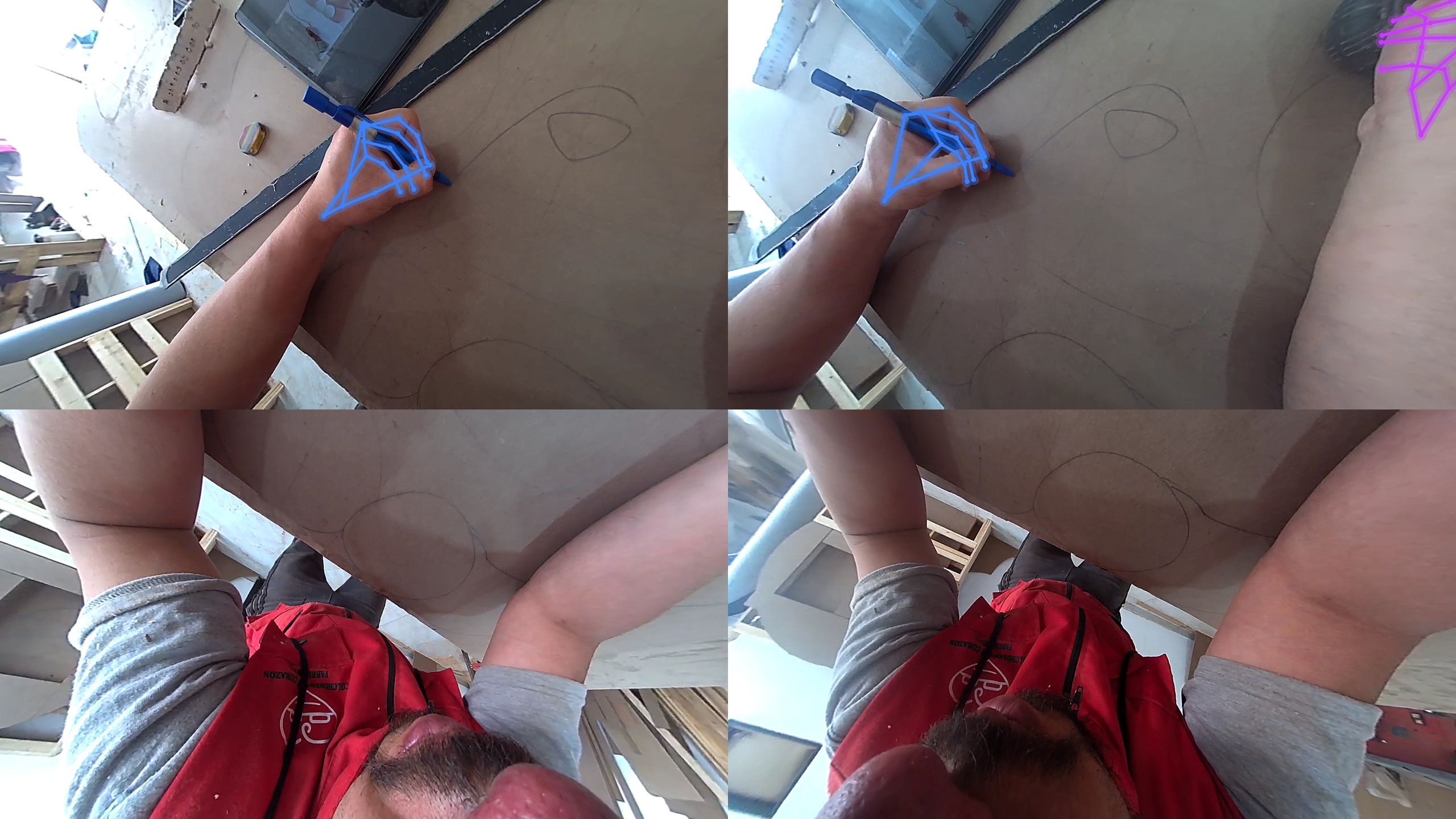} &
    \includegraphics[width=0.46\linewidth]{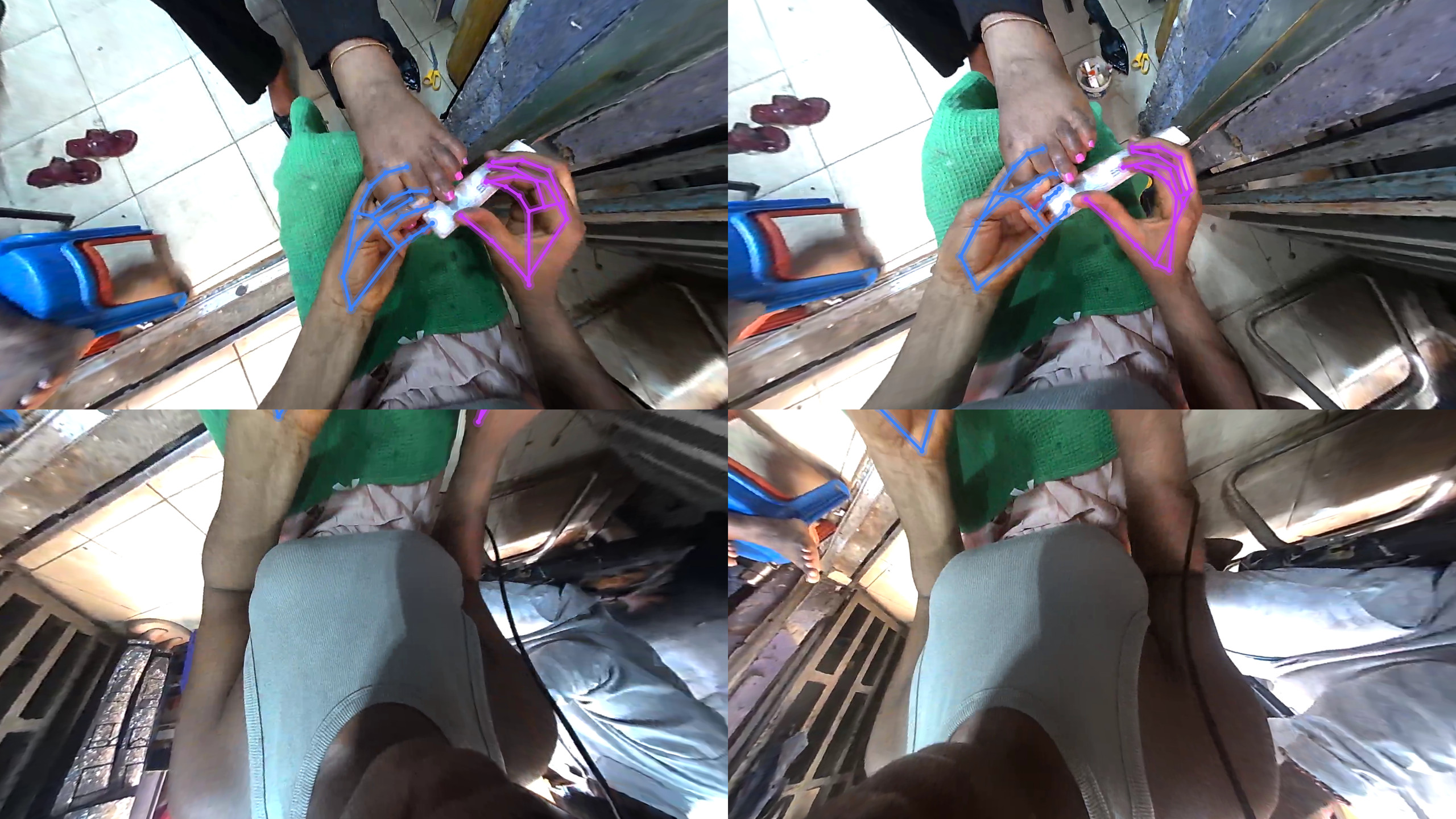} \\
    \small (e) Drawing &
    \small (f) Feet and Hands \\
  \end{tabular*}
  \caption{Robustness to hand-object interaction and unusual configurations.
  Each panel shows the four camera views of a single frame with the predicted
  mesh overlaid. The model recovers plausible poses while grasping small
  objects, while holding a tool with the fingers largely occluded by it, during
  fine manipulation such as drawing, and when the hands appear alongside other
  body parts.}
  \label{fig:hand_tracking_hard3}
\end{figure*}

We show additional qualitative results on held-out in-the-wild recordings,
covering the cases that are hardest for egocentric capture.
Figure~\ref{fig:hand_tracking_hard2} shows degraded imaging and extreme
viewpoints, and Figure~\ref{fig:hand_tracking_hard3} shows hand-object
interaction and unusual hand configurations. In each case the model produces a
plausible, multi-view consistent mesh without per-view post-processing.

\subsection{Evaluation Setup}
\label{sec:eval_setup}

\paragraph{Metric}
We report mean per-joint position error (MPJPE) over the 21 MANO joints in
millimeters. Following~\cite{poemv2}, error is absolute: it is measured in the
frame of the first camera and includes global translation, with no root
alignment and no Procrustes alignment. Camera intrinsics and extrinsics are
assumed known at both training and evaluation time. 

\subsection{Egocentric datasets}

\begin{figure*}[t]
  \centering
  \renewcommand{\arraystretch}{1.1}
  \setlength{\tabcolsep}{6pt}
  \begin{tabular}{cc}
    \includegraphics[height=0.45\linewidth]{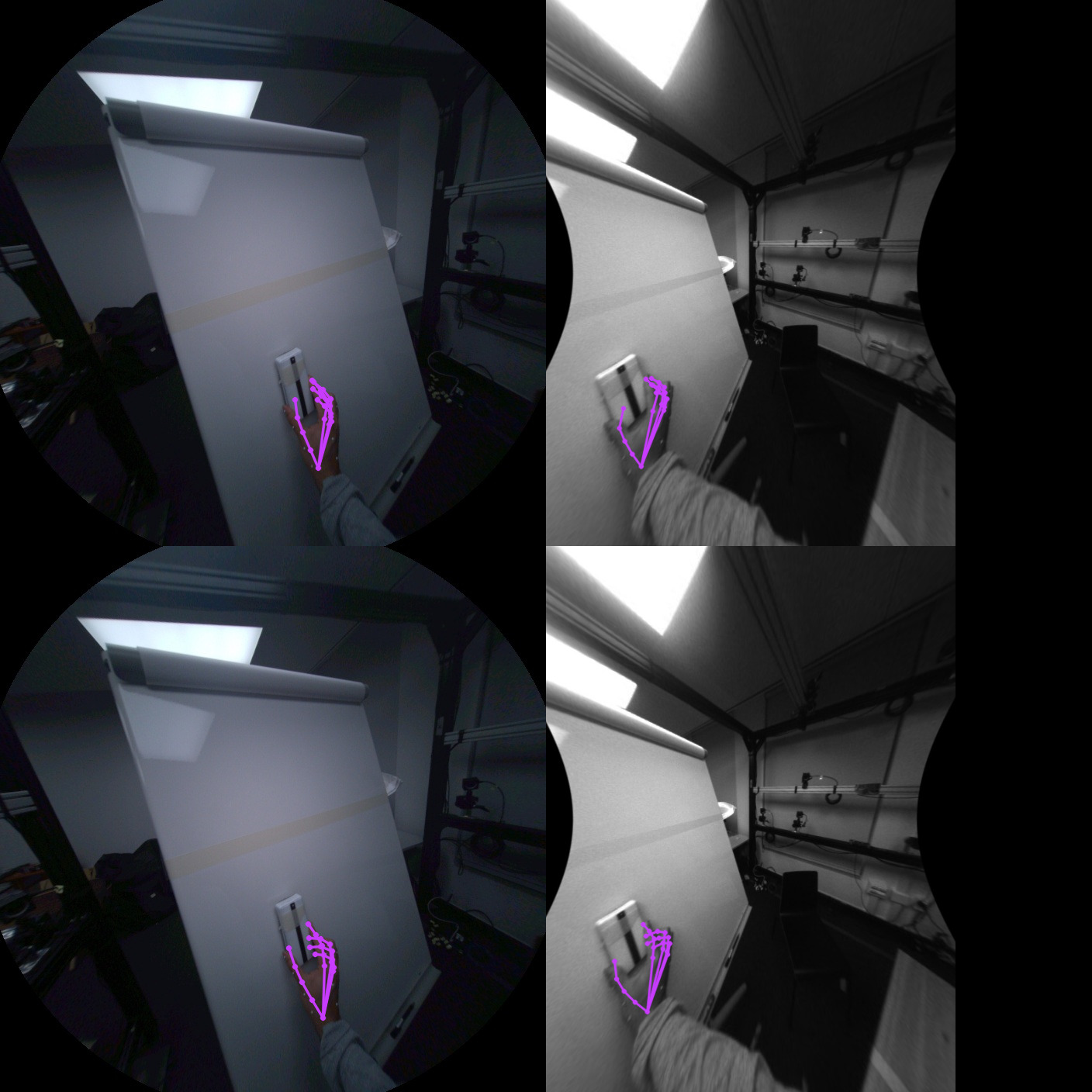} &
    \includegraphics[height=0.45\linewidth]{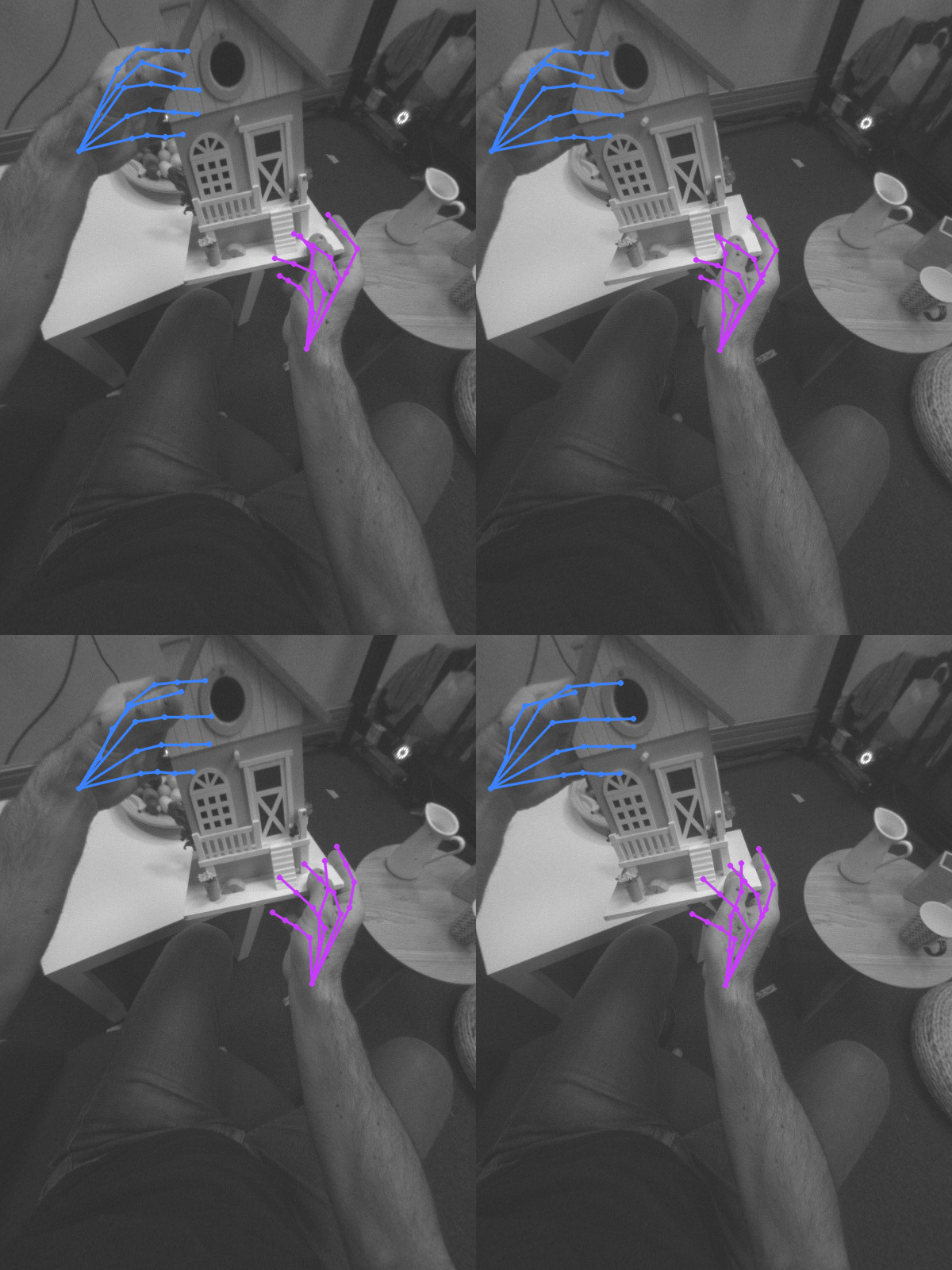} \\
    \small (a) HOT3D (Aria) &
    \small (b) HOT3D (Quest 3) \\[6pt]
    \includegraphics[height=0.45\linewidth]{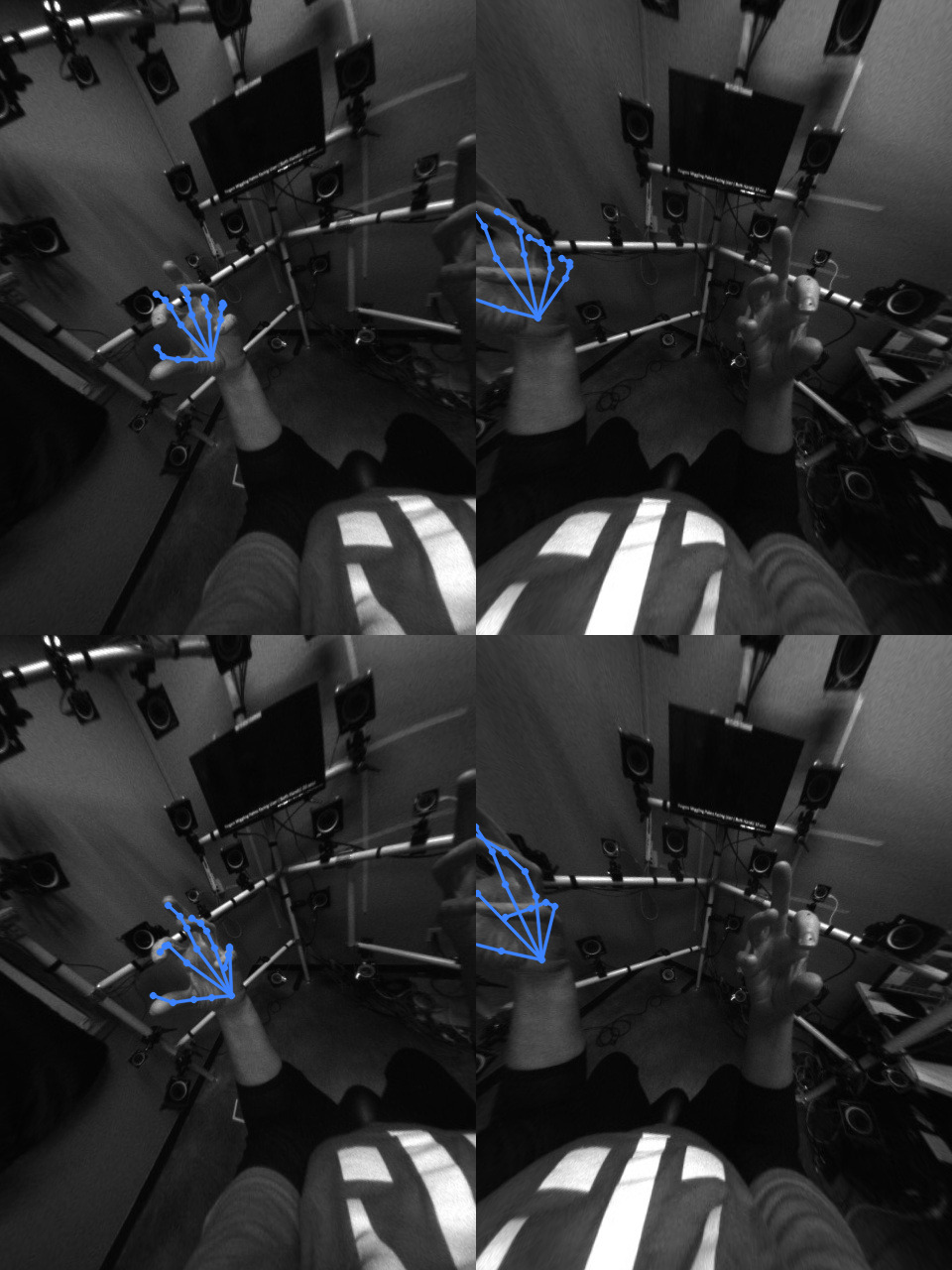} &
    \includegraphics[height=0.45\linewidth]{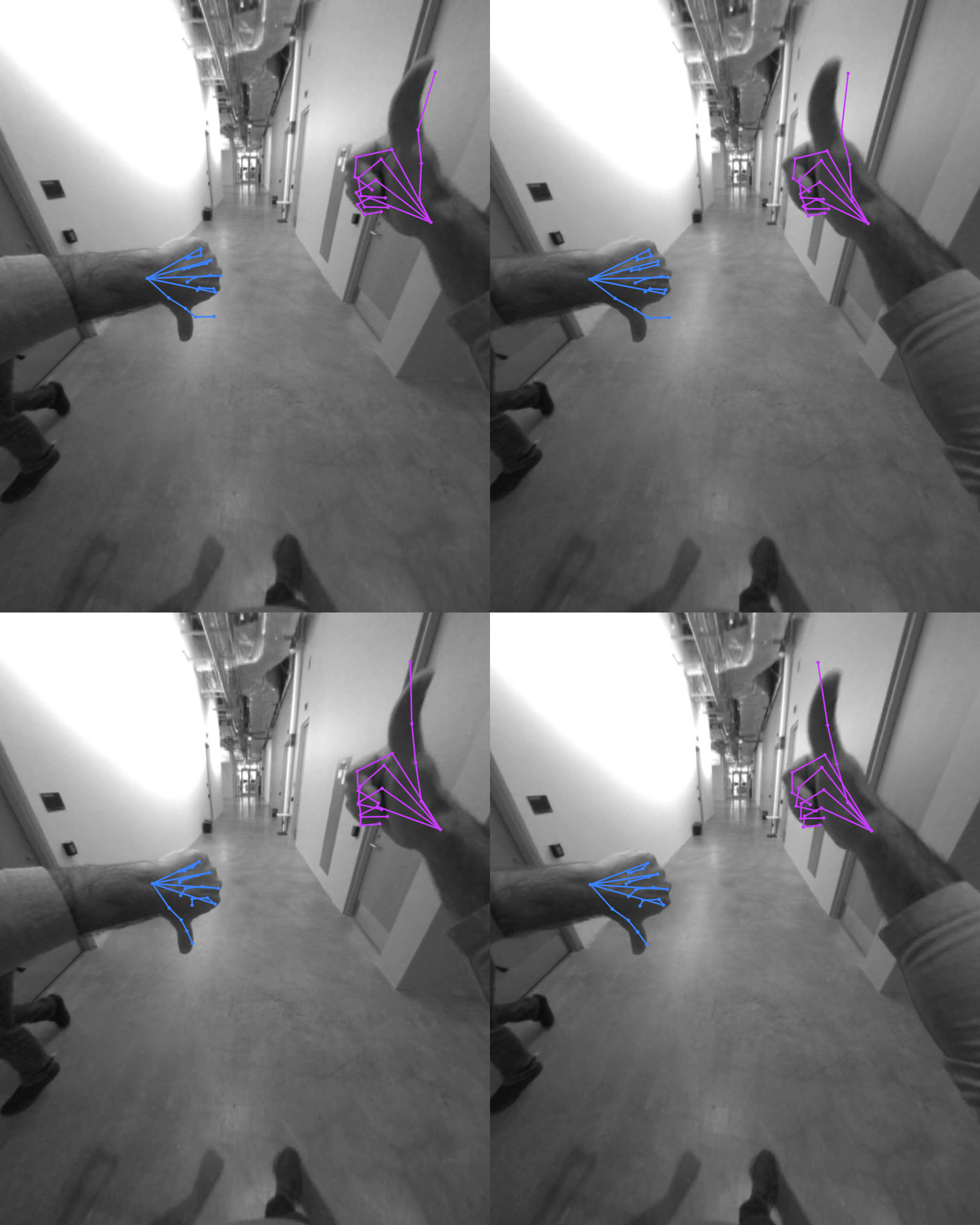} \\
    \small (c) UmeTrack &
    \small (d) SHOW3D \\
  \end{tabular}
  \caption{Comparison of random vs pretrained weights on egocentric benchmarks. Within each panel the columns are the left and right camera views of a single frame, the top row is the model trained from scratch and the bottom row is the model initialized from our pretrained weights; both are trained until they achieve their best multi-dataset validation metrics. The pretrained model aligns more closely with the observed hand, particularly at the finger joints, where the scratch model tends to mispredict articulation while keeping the global hand position roughly
  correct.}
  \label{fig:appendix_pretraining}
\end{figure*}

All three benchmarks UmeTrack~\cite{han2022umetrack}, HOT3D~\cite{banerjee2024hot3d} or
SHOW3D~\cite{rim2026show3dcapturingscenes3d} withhold the test set from public
release entirely. When evaluating on these benchmarks, we train with a 5x lower learning rate than during pretraining. The two columns of
Table~\ref{tab:ego_ablation} differ only in weight initialization.  Figure~\ref{fig:appendix_pretraining} shows sample predictions from the scratch
and pretrained models on the same frames, for each of the egocentric
benchmarks.

\paragraph{UmeTrack}
UmeTrack is a real-time hand tracking system and accompanying dataset captured on a VR headset carrying four wide-FOV monochrome VGA cameras, of which a given hand is typically visible in one or two. It targets absolute world-space pose rather than root-relative pose. The corpus comprises 53 participants and 1,397 real sequences of 15 seconds at 30\,Hz, each paired with a synthetic render, for roughly 839k frames of either kind. Ground truth comes from a 36-camera motion-capture system using 3\,mm markers, under two protocols: separate-hand and hand-hand interaction. The public release contains 41 subjects and 2,041 sequences. We hold out four subjects, matching the official pose track. Our held-out set is real-only, since the official splits contain no synthetic sequences. No subject performs both activity types, so the interaction ratio is fixed entirely by subject choice; three separate-hand subjects plus one hand-hand subject is the only composition that approaches the official ratio, giving 103 sequences at 12.6\% hand-hand against the official 95 sequences at 11.6\%. Because synthetic renders share no sequence identifier with their real counterparts, the split keys on subject, and the held-out subjects' renders are discarded rather than trained on.

\paragraph{HOT3D}
HOT3D is an egocentric dataset for 3D hand and object tracking, comprising 833 minutes of recording across 19 participants, 33 objects and 425 sessions, for over 1.5M multi-view frames and 3.7M images. It is captured on two devices: Project Aria, contributing one $1408 \times 1408$ RGB fisheye camera at $110^\circ$ field of view and two $640 \times 480$ monochrome fisheye cameras at $150^\circ$; and Quest 3, contributing two front-facing $1280 \times 1024$ monochrome fisheye cameras. Hand ground truth is obtained from several dozen OptiTrack cameras tracking 19 markers per hand, fitted to a per-participant hand model and released in both UmeTrack and MANO parameterisations. The public release covers 13 subjects across the two devices. We hold out two subjects, matching the official pose track. We select the held-out subjects jointly across both devices rather than independently per device, as selecting the Aria and Quest\,3 held-out sets independently would place the same person in training and validation. The resulting held-out set is 44 sessions and 132{,}944 frames, spanning 21 Aria and 23 Quest\,3 sessions.

\paragraph{SHOW3D}
SHOW3D is a marker-less capture dataset combining egocentric and exocentric
views: ten synchronized monochrome fisheye cameras at 60\,Hz, of which eight are
exocentric $1024 \times 1280$ cameras at $152^\circ \times 116^\circ$ field of
view mounted on a wearable rig, and two are egocentric cameras from a user-worn
Quest\,3. It comprises roughly 20 hours of recording over 38 participants, more
than 1{,}200 sequences and more than 30 locations, for 4.3M frames and 43M
images. Ground truth is produced without markers, by triangulating multi-view 2D
keypoint predictions and fitting a personalized linear-blend-skinning mesh by
inverse kinematics, with a reported sub-centimeter median accuracy. The public
release contains 32 subjects and 1{,}689 scenes. We attempt to mimic the private test set by holding out six subjects, subject to the
constraints that every object class remains represented in the training pool and
that both sides retain sufficient object-pose coverage. The resulting held-out
set is 337 scenes and 581{,}795 frames at 20.0\% of the corpus, covering 23 of
28 object classes, with a novel-activity rate of 12.7\% against the official
split's 12.2\%. SHOW3D supplies no MANO annotations, so its supervision and
evaluation are joints-only.
\begin{table}[t]
\centering
\small
\begin{tabular}{lcccc}
\toprule
& \multicolumn{2}{c}{Held-out (ours)} & \multicolumn{2}{c}{Official test} \\
\cmidrule(lr){2-3}\cmidrule(lr){4-5}
& subjects & share & subjects & share \\
\midrule
UmeTrack & 4 & 10.1\% & 4 & --- \\
SHOW3D & 6 & 20.0\% & 6 & 21.0\% \\
HOT3D & 2 & 16.0\% & 2 & --- \\
\bottomrule
\end{tabular}
\caption{Composition of the surrogate held-out sets. UmeTrack: 103 sequences,
12.6\% hand-hand interaction against the official split's 11.6\%; real
sequences only. SHOW3D: 337 scenes covering 23 of 28 object classes against the
official split's 23; novel-activity rate 12.7\% against 12.2\%. HOT3D: two
held-out subjects per device, matching the official pose track.}
\label{tab:surrogate_splits}
\end{table}

\subsection{Multi-view benchmarks}

We follow the protocol of POEMv2~\cite{poemv2} exactly, including its
multi-view subsets. Following its protocol, we only train on the right hands within each dataset. The number and ordering of cameras is fixed per dataset.

\begin{table}[t]
\centering
\small
\begin{tabular}{lcccl}
\toprule
& views & train & test & split \\
\midrule
DexYCB-Mv~\cite{chao2021dexycbbenchmarkcapturinghand} & 8 & 25{,}387 & 4{,}951 & S0 \\
OakInk-Mv~\cite{yang2022oakinklargescaleknowledgerepository} & 4 & 58{,}692 & 19{,}909 & SP2 \\
HO3D-Mv~\cite{hampali2021ho3dv3improvingaccuracyhandobject} & 5 & 9{,}087 & 2{,}706 & 7 seq. \\
\bottomrule
\end{tabular}
\caption{Multi-view benchmark splits, in multi-view frames, following~\cite{poemv2}. HO3D-Mv comprises the seven sequences: ABF1, BB1, GSF1, MDF1 and SiBF1 for training, GPMF1 and SB1 for evaluation.}
\label{tab:mv_splits}
\end{table}

A single checkpoint is finetuned for 20k steps on the union of the three
training splits and evaluated on all three test splits, with no per-dataset
specialisation. This matches POEMv2, which trains one model on a mixture and
evaluates it on each test set in turn; POEMv2's mixture additionally includes
ARCTIC, InterHand2.6M and FreiHAND, and is trained for 2.1M iterations. As
above, we reuse the pretraining configuration with the learning rate reduced
$5\times$.

\paragraph{DexYCB}
DexYCB was built to benchmark hand grasping of objects. It is captured by eight
statically mounted, calibrated RGB-D cameras at $640 \times 480$ and 30\,Hz, and
comprises 10 subjects manipulating 20 YCB objects over 1,000 sequences, for
582k frames. We adopt the official S0 split and the eight-camera construction
of~\cite{poemv2}, giving 25{,}387 training, 1{,}412 validation and 4{,}951 test
multi-view frames.

\paragraph{OakInk}
OakInk is a knowledge repository for hand-object interaction. Its image
component is captured by four calibrated RGB cameras at $848 \times 480$ and
30\,Hz, covering 12 subjects and 100 objects over roughly 230k images. We adopt
the official SP2 object-based split and the four-camera construction
of~\cite{poemv2}, giving 58{,}692 training and 19{,}909 test multi-view frames.
Roughly a quarter of OakInk sequences record two people handing an object over,
placing both hands in frame; each hand is trained and evaluated independently.

\paragraph{HO3D}
HO3D provides 3D hand and object pose annotations for hand-object manipulation,
obtained by optimising against 2D keypoint and segmentation cues rather than
from markers; v3 refines the annotation accuracy of v2. It covers 10 subjects
and 10 YCB objects over 68 sequences and roughly 103k RGB-D frames at
$640 \times 480$, of which only a subset is recorded from multiple cameras. We
adopt the construction of~\cite{poemv2}, which retains the seven sequences with
five-camera coverage: ABF1, BB1, GSF1, MDF1 and SiBF1 for training, and GPMF1
and SB1 for evaluation, giving 9{,}087 training and 2{,}706 test multi-view
frames. Two of the five cameras are withheld from the SB1 evaluation sequence.
Since absolute error is measured in the frame of the first camera, the missing
views place the reference origin on a different physical camera, approximately
$0.35$\,m from its position in every training sequence.

\subsection{Synthetic Dataset}
\label{sec:appendix_synthetic_data}

We render a synthetic dataset using the camera geometry of our RoboCap devices, which yields exact ground-truth MANO parameters. Each sequence is rendered through a camera rig whose intrinsics and extrinsics are sampled from the calibrations of real RoboCap devices, so that the synthetic imagery reproduces the field of view, baseline and lens distortion of our hardware.


\end{document}